\documentclass[sigconf]{acmart}
\usepackage{amsfonts}       % blackboard math symbols
\usepackage{amsmath}
\usepackage{amssymb}
\usepackage{nicefrac} % compact symbols for 1/2, etc.
\usepackage[ruled, vlined, linesnumbered]{algorithm2e}
\usepackage{enumerate}
\usepackage{enumitem}
\usepackage{geometry}
\usepackage{tikz}
\usetikzlibrary{shapes.geometric}
\usepackage{arydshln}
\usepackage{xspace}
\usepackage{amssymb}
\usepackage{bbding}
\usepackage{tcolorbox}
\usepackage{multirow}
\usepackage{threeparttable}
\usepackage{soul}
\usepackage{color}

\usepackage[capitalize,noabbrev]{cleveref}
\usepackage{makecell}
\usepackage{wrapfig}
\usepackage{caption}
\usepackage{subcaption}
\usepackage{pifont}
\usepackage{tikz}
\usepackage{hyperref}
\usepackage[thicklines]{cancel}
\SetKwInput{KwInput}{Input} 
\SetKwInput{KwOutput}{Output}  
\usepackage{graphicx} 
\usepackage[export]{adjustbox}
\usepackage{balance}
\usepackage[table]{xcolor}
\usepackage{booktabs}
\usepackage{arydshln}
\definecolor{lightpurple}{HTML}{c6c6ff}

\newcommand{\dlight}[1]{\cellcolor{red!10}#1}
\newcommand{\dmid}[1]{\cellcolor{red!25}#1}
\newcommand{\ddeep}[1]{\cellcolor{red!45}\textbf{#1}}

\newcommand{\z}{\textcolor{gray!60}{0.00000}}

\setulcolor{red}

\AtBeginDocument{%
  \providecommand\BibTeX{{%
    \normalfont B\kern-0.5em{\scshape i\kern-0.25em b}\kern-0.8em\TeX}}}

\setcopyright{acmcopyright}
\copyrightyear{2018}
\acmYear{2018}
\acmDOI{XXXXXXX.XXXXXXX}

\acmConference[Conference acronym 'XX]{Make sure to enter the correct
  conference title from your rights confirmation email}{June 03--05,
  2018}{Woodstock, NY}
\acmPrice{15.00}
\acmISBN{978-1-4503-XXXX-X/18/06}

\begin{document}
% \title{Cold-Item Reachability in Generative Recommendation: A Temporal Study of Semantic-ID Generation}
\title{Can Generative Recommendation Reach Cold Items? A Temporal Perspective on Semantic-ID Generation}

%%
%% The "author" command and its associated commands are used to define the authors and their affiliations.
\author{Jie Peng}
\affiliation{%
  \institution{Renmin University of China}
  \city{Beijing}
  \country{China}
}
\email{peng_jie@ruc.edu.cn}

\author{Yanping Zheng}
\affiliation{%
  \institution{Renmin University of China}
  \city{Beijing}
  \country{China}
}
\email{zhengyanping@ruc.edu.cn}

\author{Zhewei Wei}
\affiliation{%
  \institution{Renmin University of China}
  \city{Beijing}
  \country{China}
}
\authornote{Corresponding Authors.}
\email{zhewei@ruc.edu.cn}

\author{{Bin Tong}}
\affiliation{%
  \institution{Alibaba}
  \city{Beijing}
  \country{China}
}
\email{tongbin.tb@alibaba-inc.com}

\author{Guan Wang}
\affiliation{%
  \institution{Alibaba}
  \city{Beijing}
  \country{China}
}
\email{shangfeng.wg@taobao.com}

\author{Bo Zheng}
\affiliation{%
  \institution{Alibaba}
  \city{Beijing}
  \country{China}
}
\email{bozheng@alibaba-inc.com}

% \author{}
% \affiliation{%
%   \institution{Alibaba}
%   %\city{Beijing}
%  % \country{China}
% }
% \email{}

\begin{abstract}
Semantic-ID-based generative recommendation represents items as sequences of shared semantic tokens, enabling token recombination beyond isolated item IDs. However, closed-world recombination does not necessarily imply temporal open-token cold-start induction, where new items enter the item catalog with unseen atomic tokens or weakly supported SID paths. In this work, we revisit SID-based generative recommendation under an absolute-time temporal protocol that separates seen and unseen targets and diagnoses the cold item reachability at the token level. Through seen/unseen-hit analysis, coldness taxonomy, and oracle-prefix probing, we show that current SID-based models can occasionally reach future items supported by observed tokens and prefixes, but struggle with unseen atomic tokens and unsupported SID paths. We further explain this boundary by interpreting SID generation as hierarchical semantic bucketing: early tokens select coarse semantic regions, while later tokens refine item-specific paths. These findings show that SID generation is compositional but not fully open-ended, and suggest future directions in more independent SID spaces, scoring-based interfaces, and dynamic textual context.

\end{abstract}

% openreview tex version

%    <concept>
%        <concept_id></concept_id>
%        <concept_desc></concept_desc>
%        <concept_significance>500</concept_significance>
%        </concept>
%  </ccs2012>

% \ccsdesc[500]{}
% \vskip -1in
% \setlength{\parskip}{0pt}
\vspace{-10pt}

\keywords{Generative Recommendation, Semantic ID, Cold Item Reachability, Token-level Cold-start}
\vspace{-10pt}

\maketitle

\section{Introduction}\label{intro}
\begin{figure}[t]
\centering
% \vskip -0.1in
\includegraphics[width=0.4\textwidth]{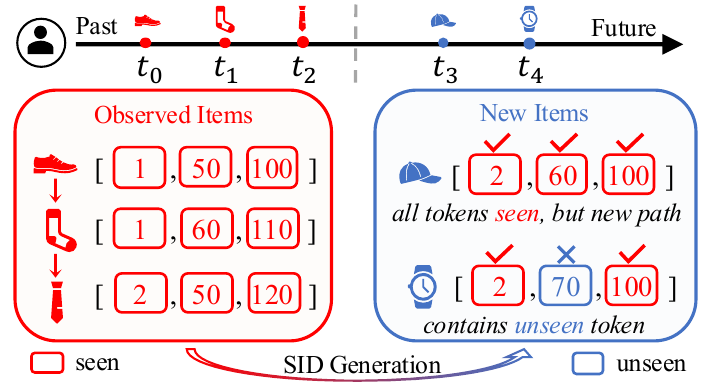}
\vskip -0.1in
\caption{Temporal cold start as token-level reachability.}
\vskip -0.2in
\label{fig:intro}
\end{figure}

Recommendation systems have traditionally represented items with randomly assigned or hashed identifiers \cite{koren2009matrix,cheng2016wide,zhao2019recommending}.
While effective for memorizing historical user--item interactions, such ID-based representations treat items as isolated symbols and offer limited structural sharing across semantically related items.
This limitation becomes more pronounced in large-scale recommendation, where catalogs are long-tailed, sparse, and continuously evolving \cite{singh2024better}.
Semantic IDs (SIDs) provide an alternative by representing each item as a sequence of discrete semantic tokens derived from content or representation spaces \cite{tay2022transformer,rajput2023recommender,singh2024better}.
By sharing tokens or prefixes across related items, SIDs enable semantic-aware modeling and compactly encode large catalogs with reusable codewords.

\begin{figure*}[t]
\centering
\vskip -0.05in
\includegraphics[width=0.9\textwidth]{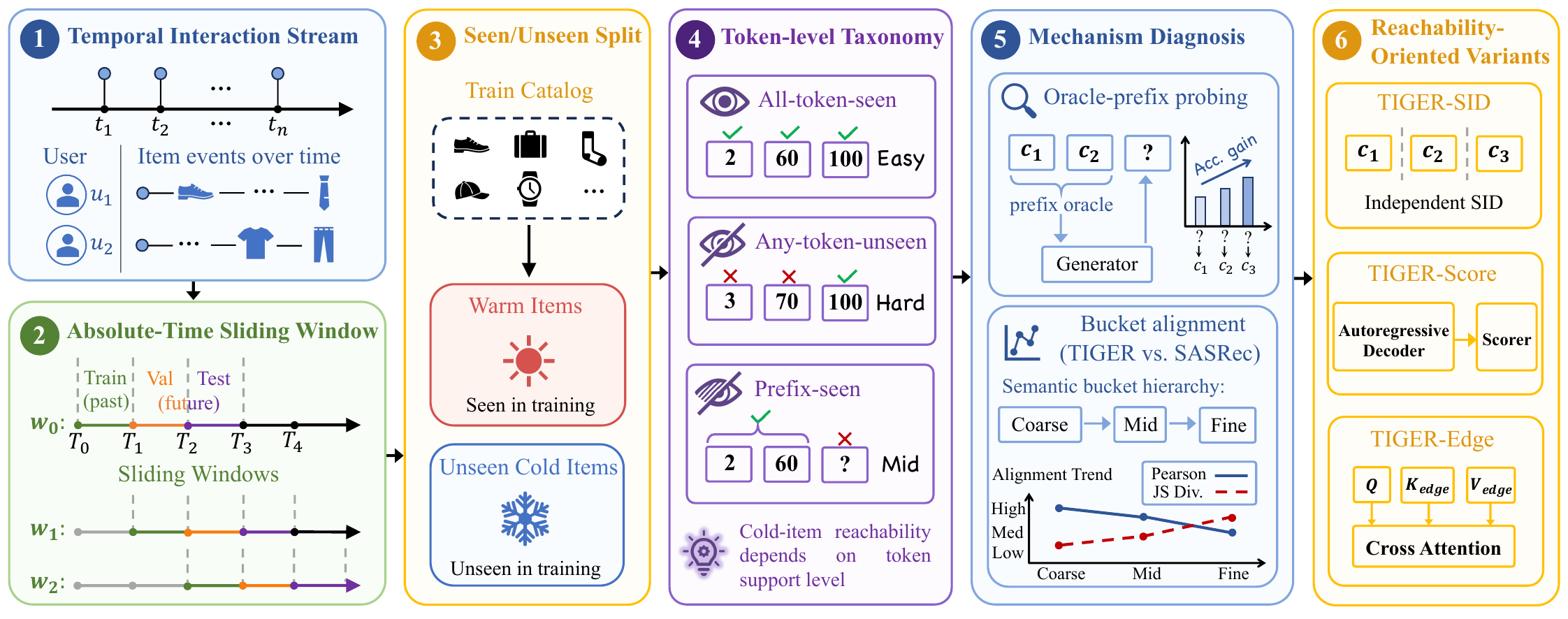}
\vskip -0.15in
\caption{Overview of temporal cold-start reachability analysis.}
\vskip -0.15in
\label{fig:pipeline}
\end{figure*}

SID-based generative recommendation (GR) formulates next-item prediction as autoregressive generation of the target item's SID, rather than scoring over items \cite{rajput2023recommender}.
This enables \emph{closed-token recombination}: a model may compose observed tokens or prefixes into SID paths that were never seen as complete item targets during training.
However, such recombination is not equivalent to cold-start induction.
Real systems must recommend newly introduced items with limited or no interaction history \cite{schein2002methods,lam2008addressing}.
Some future items can be expressed using familiar SID tokens, while others require unsupported paths or atomic tokens never observed as training targets.
This raises the central question of this paper: \emph{can SID-based GR inductively reach future cold items, or is its reachability bounded by the item--token support observed during training?}

Answering this question requires rethinking evaluation first.
The standard leave-one-out protocol supports closed-world sequential prediction \cite{kang2018self,sun2019bert4rec}, but splits each user sequence independently and may interleave train, validation, and test interactions in absolute time.
Thus, a future test item for one user may already appear in another user's training history, obscuring the difficulty of recommending items that emerge only after training.
Manually removing items can simulate cold start, but still differs from the natural online setting where items arrive along a global timeline \cite{rajput2023recommender,zhang2026cold}.

We therefore introduce an absolute-time temporal cold-start protocol based on sliding temporal windows.
In each window, models are trained on past interactions and evaluated on future interactions, so cold test items correspond to items emerging after training.
This protocol exposes diverse temporal regimes, yields richer cold-item and SID-token distributions, explicitly separates seen and unseen targets, and directly measures whether SID-based GR can reach beyond the item and token support available during training.

Under this protocol, we re-evaluate TIGER \cite{rajput2023recommender}, a representative SID-based generative recommender, on multiple datasets \cite{mcauley2015image,peng2025gdgb}.
Compared with leave-one-out evaluation, TIGER suffers substantial degradation in Recall and NDCG, showing that closed-world strength does not directly translate into open-world reachability.
To locate the source of this degradation, we diagnose cold items at the token level.
Beyond the binary seen/unseen distinction, we refine cold-start difficulty by SID support: \textit{all-token-seen} items whose SID tokens all appeared during training, \textit{any-token-unseen} items containing at least one unseen atomic token, and \textit{prefix-seen} items whose SID prefix was observed but whose full path may be new.
We further perform oracle-prefix probing, where the correct first token or first two tokens are given to the decoder, separating coarse bucket-selection errors from fine-grained path-completion failures.

As illustrated in \cref{fig:intro}, our analysis reveals a structural reachability boundary.
TIGER can occasionally generate cold-item SIDs when the required tokens and prefixes are sufficiently supported by training interactions, but struggles with unseen atomic tokens or unsupported SID paths.
Successful unseen-item hits concentrate in token- and prefix-supported regimes, whereas unsupported cases are substantially harder to reach.
Thus, SID-based generation is not open-ended item generation in the same sense as natural language generation; it is a constrained search over a learned discrete token-path space.
Its cold-start capability is compositional, but not fully expansive under current paradigm.

To explain this boundary, we reinterpret SID generation as coarse-to-fine semantic bucket classification, akin to that in SASRec~\citep{kang2018self}.
By aligning TIGER's token-generation distribution with SASRec's bucket-aggregated scoring distribution, we find that early SID positions mainly capture coarse semantic preferences, while later positions perform increasingly fine-grained path refinement.
This explains both the promise and limitation of SID-based GR: semantic compression and token sharing organize items into reusable buckets and paths, but inductive reach remains bounded by the learned token vocabulary and path support.

Finally, we study three TIGER-based diagnostic variants that intervene on different reachability bottlenecks.
\textbf{TIGER-SID} modifies the SID code space with a more factorized tokenizer while keeping autoregressive generation unchanged \cite{ge2013optimized,hou2025generating};
\textbf{TIGER-Scorer} keeps the TIGER encoder but replaces exact SID decoding with candidate scoring, isolating user modeling from SID generation;
and \textbf{TIGER-Edge} preserves the generative SID objective while injecting dynamic edge context through cross attention.
Together, these variants probe whether cold-item reachability can be expanded through SID design, scorer interfaces, and contextual signals.

In summary, as shown in \cref{fig:pipeline}, this work clarifies the cold-item reachability boundary of SID-based generative recommendation under temporal cold start.
Our contributions are as follows:
\begin{itemize}
\item We propose an absolute-time sliding-window protocol to fairly evaluate SID-based GR at the item and token levels.
\item We use a token-level coldness taxonomy and oracle-prefix probing to diagnose cold-item reachability bottlenecks.
\item We show that SID generation behaves as hierarchical bucket classification and explore three TIGER-based variants for expanding reachability.
\end{itemize}

\section{Related Work and Positioning} \label{background}

\begin{figure*}[t]
\centering
\vskip -0.2in
\includegraphics[width=0.92\textwidth]{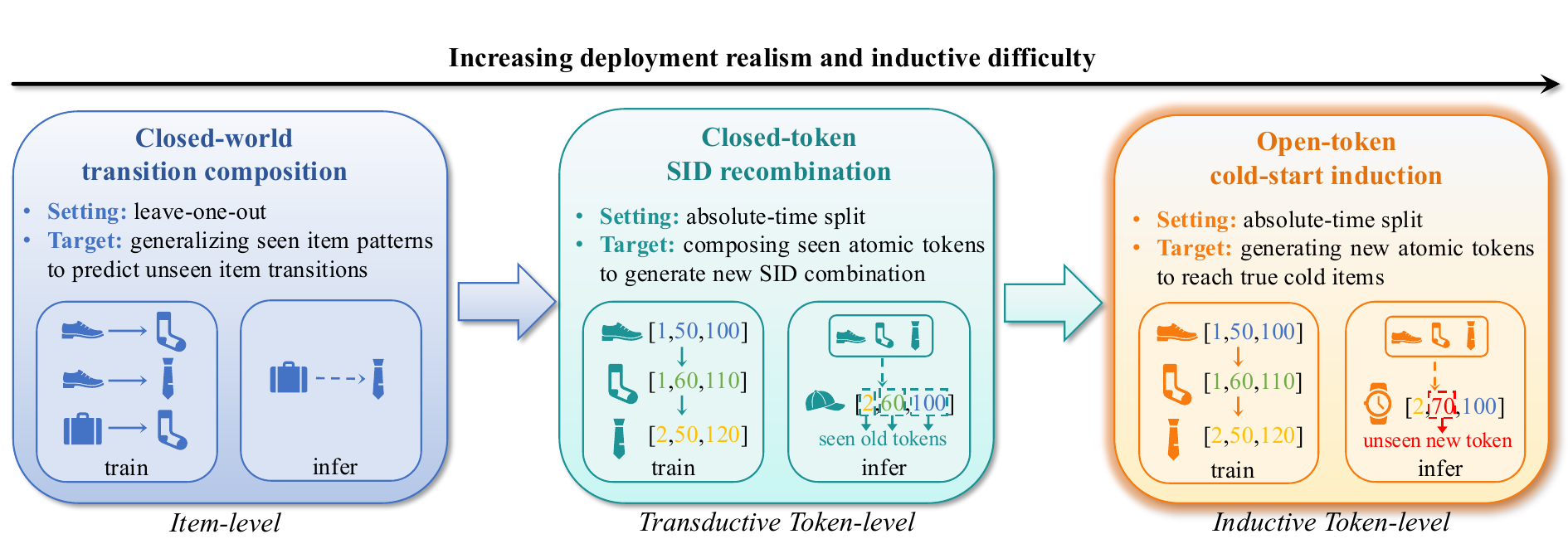}
\vskip -0.1in
\caption{Regime spectrum of deployment realism and inductive difficulty.}
\vskip -0.15in
\label{fig:induction_levels}
\end{figure*}

\subsection{From ID-based Recommendation to Semantic-ID Generation}

Classical recommender systems represent each item with an independent identifier, such as a one-hot ID, learned embedding, or hashed feature.
This design supports a wide range of collaborative filtering and neural recommendation models, including matrix factorization, wide-and-deep models, and Transformer-based sequential recommenders \citep{koren2009matrix,cheng2016wide,kang2018self,sun2019bert4rec}.
While effective in warm-start settings, ID-based representations offer limited structural sharing and therefore struggle in long-tailed or evolving catalogs with new items \citep{schein2002methods,lam2008addressing}.
SIDs address this limitation by mapping each item into a sequence of discrete semantic codes,
\begin{equation}
\mathbf{c}(i) = (c_1(i), c_2(i), \ldots, c_L(i)),
\end{equation}
typically obtained from item content embeddings or learned representation spaces through vector quantization.
This connects SID construction to discrete representation learning and quantization methods, including VQ-VAE, residual quantization (RQ), product quantization (PQ), and optimized product quantization (OPQ) \citep{oord2017vqvae,lee2022rqvae,jegou2011pq,ge2013opq}.
Because related items may share tokens or prefixes, SIDs allow recommenders to reuse statistical strength across items.

SID-based generative recommendation further casts next-item prediction as autoregressive generation of the target SID:
\begin{equation} p(i_{t+1} \mid \mathcal{H}_u) \approx \prod_{\ell=1}^{L} p\!\left(c_\ell(i_{t+1}) \mid \mathcal{H}_u, c_{<\ell}(i_{t+1})\right), \end{equation}
where $\mathcal{H}_u$ denotes the historical interactions of user $u$.
TIGER is a representative framework in this direction \citep{rajput2023recommender}, while broader generative recommendation also connects to language-model-based frameworks such as P5 \citep{geng2022p5}.
Recent work further shows that SID construction is crucial: CoST improves semantic tokenization through contrastive quantization, and LETTER learns item tokenizers with semantic and collaborative regularization \citep{zhu2024cost,wang2024letter,hou2025actionpiece}.
These studies establish SID quality as a central factor in SID-based GR.

\subsection{Closed-World Composition in GR}

A key promise of SID-based GR is token sharing and recombination.
Since an item is decomposed into multiple semantic tokens, the model may compose observed tokens or prefixes into new SID sequences.
This differs from conventional ID-based scoring, where each item is usually ranked as an independent candidate.
Recent studies therefore analyze whether GR gains arise from memorization, semantic sharing, or compositional transition patterns.

A relevant line of work studies generative recommendation under closed-world sequential prediction.
For example, recent analyses categorize test cases according to whether their targets can be solved by memorizing observed item transitions or composing known transition patterns \citep{ding2026well}.
Such work shows that apparent item-level novelty may often be explained by token-level memorization or prefix sharing in SID space, helping explain why SID-based GR can perform well under leave-one-out evaluation.

Our work is complementary but focuses on a different regime.
Prior analyses mainly study closed-world transition interpolation with a largely fixed item universe.
We study temporal open-token cold-start induction, where new items, SID paths, and possibly atomic tokens emerge after training.
Thus, we do not question the closed-world compositionality of GR; instead, we ask where it stops being sufficient under realistic temporal deployment:
\begin{quote}
\emph{When the item catalog evolves over time, which cold items can SID-based GR reach, which cold items remain outside its token-supported generation space, and why?}
\end{quote}

\subsection{Cold-start under Temporal Deployment}

Cold-start recommendation is a long-standing challenge in recommender systems \citep{schein2002methods,lam2008addressing}.
A common solution is to incorporate side information, such as metadata, text, images, categories, or graph structure, so that new items can be represented before sufficient collaborative signals accumulate.
Neural and multimodal methods further combine collaborative and content signals, for example through dropout-based training for missing collaborative information or multimedia representations for new items \citep{volkovs2017dropoutnet,wei2017coldstart,bai2024milk}.
These approaches emphasize that inductive recommendation generally requires information beyond historical item IDs.

In sequential recommendation, however, cold-start evaluation is often blurred by leave-one-out protocols.
Although leave-one-out preserves each user's interaction order, it does not preserve the global chronological order of interactions.
A test item for one user may already appear in other users' training histories, leading to optimistic estimates of deployment-time cold-start ability.
Recent studies show that splitting strategies substantially affect reported model rankings and that ignoring global time can introduce temporal leakage \citep{meng2020splitting,ji2020leakage,gusak2025timetosplit,peng2025tidformer,peng2025beyond}.
Absolute-time temporal splits are thus more appropriate for deployment-oriented cold-start analysis.

This issue is especially important for SID-based GRs.
A future item may be reachable if its SID can be composed from observed tokens and prefixes.
However, if it requires unseen atomic tokens or unsupported SID paths, the autoregressive decoder may have no reliable way to generate it.
Thus, the relevant cold-start question is not only whether the target item ID has been seen, but whether its token-level support is available.

\subsection{Positioning of This Work}

We position this paper as a diagnostic study of SID-based generative recommendation under temporal cold start.
As shown in \cref{fig:induction_levels}, prior work mainly studies ID-based recommendation, SID-based closed-world generation, or closed-world recombination under leave-one-out protocols.
In contrast, we focus on a deployment-oriented regime where items, SID paths, and token support evolve under absolute-time splits.

Our goal is not to challenge the closed-world compositional benefit of GRs, but to characterize its reachability boundary under absolute-time cold-start.
In this paper, we combine temporal evaluation, token-level coldness taxonomy, oracle-prefix probing, and bucket-level comparison with discriminative recommendation models, and further use controlled TIGER-based variants to examine whether this boundary can be expanded through SID-space design, scoring-based interfaces, and dynamic contextual signals.

\begin{table*}[t]
\centering
% \vskip -0.1in
\caption{Seen and unseen target performance under temporal cold-start evaluation. ``Cold ratio'' denotes the average proportion of test targets whose item IDs are absent from the training period. Hit@20 and Recall@20 denote the average values among all windows. Detailed results for each window and other order metrics are presented in \cref{tab:tiger_seen_unseen_at5,tab:sasrec_seen_unseen_at5,tab:tiger_seen_unseen_at10,tab:sasrec_seen_unseen_at10,tab:tiger_seen_unseen_at20,tab:sasrec_seen_unseen_at20}.}
\vskip -0.1in
\label{tab:seen_unseen_main}
\resizebox{0.95\textwidth}{!}{
\begin{tabular}{l c cccc cccc}
\toprule
\multirow{2}{*}{\textbf{Dataset}}
& \multirow{2}{*}{\textbf{Cold ratio}}
& \multicolumn{4}{c}{\textbf{TIGER}}
& \multicolumn{4}{c}{\textbf{SASRec}} \\
\cmidrule(lr){3-6}\cmidrule(lr){7-10}
&
& Seen H@20 & Seen R@20 & Unseen H@20 & Unseen R@20
& Seen H@20 & Seen R@20 & Unseen H@20 & Unseen R@20 \\
\midrule
Beauty      & 0.948& 0.01334 & 0.00258 & 0.00023 & \z & \textbf{0.08367} & \textbf{0.01170} & \textbf{0.00414} & \textbf{0.00068} \\
Sports      & 0.947& 0.01406 & 0.00851 & 0.00020 & 0.00004 & \textbf{0.04968} & \textbf{0.06943} & \textbf{0.00416} & \textbf{0.00022} \\
Toys        & 0.910& 0.02340 & 0.01371 & 0.00009 & \z & \textbf{0.05232} & \textbf{0.05590} & \textbf{0.00146} & \textbf{0.00003} \\
\midrule
Sephora     & 0.773& \textbf{0.33718} & \textbf{0.33749} & \z & \z & 0.07339 & 0.08744 & \textbf{0.05665} & \textbf{0.02532} \\
WikiLife    & 0.569& \textbf{0.39112} & \textbf{0.33868} & 0.00103 & 0.00042 & 0.04037 & 0.06678 & \textbf{0.01163} & \textbf{0.01090} \\
WeiboDaily  & 0.253& \textbf{0.75894} & 0.40206 & \textbf{0.00313} & \textbf{0.00092} & 0.27167 & \textbf{0.45276} & 0.00191 & 0.00003 \\
WeiboTech   & 0.385& \textbf{0.41039} & 0.06812 & \z & \z & 0.33391 & \textbf{0.09163} & \z & \z \\
\bottomrule
\end{tabular}}
\vskip -0.1in
\end{table*}

\section{Temporal Cold-Start Evaluation}\label{temporal_coldstart}

\subsection{Temporal Interaction Stream}

Let the recommendation data be a timestamped interaction stream $\mathcal{D} =\{(u, i, t, \mathbf{x}_{ui})\}$, where $u \in \mathcal{U}$ is a user, $i \in \mathcal{I}$ is an item, $t$ is the absolute timestamp, and $\mathbf{x}_{ui}$ optionally denotes side information such as textual attributes, edge content, or interaction metadata.
For each user $u$, the historical sequence before time $t$ is
\begin{equation}
    \mathcal{H}_u^{<t}
    =
    \{(i_1,t_1), (i_2,t_2), \ldots, (i_m,t_m)\},
    \quad
    t_1 < t_2 < \cdots < t_m < t .
\end{equation}
The goal is to predict the next item $i$ that user $u$ will interact with after observing $\mathcal{H}_u^{<t}$.
Unlike leave-one-out evaluation, which splits each user sequence independently, we partition the entire interaction stream by absolute time.
For a temporal window $w$, let $\tau_w^{\mathrm{tr}} < \tau_w^{\mathrm{val}} < \tau_w^{\mathrm{te}}$ be the training, validation, and test boundaries.
The three splits are defined as
\begin{align}
    \mathcal{D}^{w}_{\mathrm{train}}
    &=
    \{(u,i,t,\mathbf{x}_{ui}) \in \mathcal{D}
    \mid t \leq \tau_w^{\mathrm{tr}}\}, \\
    \mathcal{D}^{w}_{\mathrm{val}}
    &=
    \{(u,i,t,\mathbf{x}_{ui}) \in \mathcal{D}
    \mid \tau_w^{\mathrm{tr}} < t \leq \tau_w^{\mathrm{val}}\}, \\
    \mathcal{D}^{w}_{\mathrm{test}}
    &=
    \{(u,i,t,\mathbf{x}_{ui}) \in \mathcal{D}
    \mid \tau_w^{\mathrm{val}} < t \leq \tau_w^{\mathrm{te}}\}.
\end{align}
The window is then slid forward to construct multiple temporal regimes.
This design ensures that the model is always trained on past interactions and evaluated on future interactions.

\subsection{Seen and Unseen Target Items}

For each temporal window $w$, the training item set is
$
    \mathcal{I}^{w}_{\mathrm{train}}
    =
    \{i \mid (u,i,t,\mathbf{x}_{ui}) \in \mathcal{D}^{w}_{\mathrm{train}}\}$.
A test target item $i$ is considered {seen} if it appeared in the training period: $i \in \mathcal{I}^{w}_{\mathrm{seen}}
    \Longleftrightarrow
    i \in \mathcal{I}^{w}_{\mathrm{train}}$,
and {cold} ({unseen}) otherwise: $
    i \in \mathcal{I}^{w}_{\mathrm{cold}}
    \Longleftrightarrow
    i \notin \mathcal{I}^{w}_{\mathrm{train}}$.
Accordingly, the test set is partitioned into
\begin{align}
    \mathcal{D}^{w}_{\mathrm{test,seen}}
    &=
    \{(u,i,t,\mathbf{x}_{ui}) \in \mathcal{D}^{w}_{\mathrm{test}}
    \mid i \in \mathcal{I}^{w}_{\mathrm{seen}}\}, \\
    \mathcal{D}^{w}_{\mathrm{test,cold}}
    &=
    \{(u,i,t,\mathbf{x}_{ui}) \in \mathcal{D}^{w}_{\mathrm{test}}
    \mid i \in \mathcal{I}^{w}_{\mathrm{cold}}\}.
\end{align}
The cold ratio of window $w$ is $
    \rho^{w}_{\mathrm{cold}}
    =
    \frac{
    |\mathcal{D}^{w}_{\mathrm{test,cold}}|
    }{
    |\mathcal{D}^{w}_{\mathrm{test}}|
    }$.
This quantity measures how strongly each window reflects temporal item emergence.

\subsection{SID Token Support}

For SID-based GR, item-level coldness is not sufficiently informative.
A cold item may still be expressible through tokens that were observed during training.
Let the SID of item $i$ be $\mathbf{c}(i) = (c_1(i), \ldots, c_L(i))$.
The set of atomic tokens observed in the training period is $\mathcal{V}^{w}_{\mathrm{train}}
    =
    \{c_\ell(i)
    \mid i \in \mathcal{I}^{w}_{\mathrm{train}},
    \ell \in \{1,\ldots,L\}\}$.
A token $c_\ell(i)$ is seen if $c_\ell(i)\in \mathcal{V}^{w}_{\mathrm{train}}$ and unseen otherwise.
We further define the set of observed SID prefixes:
$\mathcal{P}^{w}_{\ell}
    =
    \{(c_1(i),\ldots,c_\ell(i))
    \mid i \in \mathcal{I}^{w}_{\mathrm{train}}\}$.
A cold item can therefore be characterized not only by whether its item ID appeared in training, but also by whether its atomic tokens and prefixes appeared in training.

\subsection{Model Architecture and Evaluation}

Our primary generative model follows the standard TIGER-style architecture.
An item tokenizer maps each item to a length-$L$ SID sequence.
Given a user history $\mathcal{H}_u$, the encoder produces a sequence representation, and the decoder autoregressively predicts the target SID:
\begin{equation}
    \mathcal{L}_{\mathrm{GR}}
    =
    - \sum_{(u,i)\in \mathcal{D}_{\mathrm{train}}}
    \sum_{\ell=1}^{L}
    \log
    p_{\theta}\!\left(
    c_\ell(i)
    \mid
    \mathcal{H}_u, c_{<\ell}(i)
    \right).
\end{equation}
At inference time, beam search is used to generate candidate SID sequences.
Valid generated SIDs are mapped back to item candidates through the codebook.
We evaluate recommendation quality with Recall@$K$ and NDCG@$K$ on overall, seen-target, and cold-target test subsets.
Besides, we use Hit@$K$ to evaluate each test sequence once against its full target item set. A Hit@$K$ is counted as 1 if any of the top-$K$ predicted items matches at least one item in the corresponding target set, and 0 otherwise. 
We compare TIGER with SASRec \citep{kang2018self}, a representative discriminative sequential recommender that scores candidate items directly.
This comparison allows us to distinguish the effect of SID-based autoregressive generation from conventional item-level scoring.

% \subsection{Tokenizer Setting}

% The temporal evaluation protocol also requires clarity about the tokenizer.
% If the tokenizer is fitted on all items, including future items, the SID vocabulary may contain information unavailable at training time.
% If the tokenizer is fitted only on training-period items, it better reflects deployment but may require out-of-vocabulary handling for future items.
% In our analysis, we explicitly report the tokenizer setting used in each experiment.
% When an all-item tokenizer is used, we interpret the results as a diagnostic upper bound on SID-token availability.
% When a train-only tokenizer is used, the evaluation more directly reflects real temporal deployment.

\section{Token-Level Diagnosis of Cold-Start Boundary}\label{token_coldstart}
\subsection{Performance under Temporal Splits}

We first revisit the next-item prediction performance of TIGER under a rigorous temporal cold-start protocol. Unlike the conventional leave-one-out setting, which often allows information leakage, our absolute-time split enforces a strict boundary where test targets emerge strictly after the training period. This setup introduces a significantly more challenging cold-start scenario, particularly for evaluating the model's inductive reachability to unseen items. \cref{tab:seen_unseen_main} details the performance of TIGER and SASRec across seven datasets. Specifically, beyond the widely used datasets in GR research, namely Beauty, Sports, and Toys \cite{mcauley2015image}, we additionally incorporate recent Dynamic Text-attributed Graph (DyTAG) datasets (i.e., Sephora, WikiLife, WeiboDaily, and WeiboTech) \cite{peng2025gdgb}. These DyTAG datasets are characterized by rich node and edge texts along with explicit temporal dynamics, thereby providing high-quality data for generative recommendation \cite{peng2025gdgb}. Furthermore, all datasets were partitioned into five sliding temporal windows, resulting in three test windows. See \cref{app:dataset} for more dataset information.

\textbf{Results.}
To expose the effect of item emergence, we separately evaluate seen and unseen target items under the temporal protocol.
As shown in \cref{tab:seen_unseen_main}, unseen targets account for a substantial portion of the test set, with cold ratios ranging from 0.2533 on WeiboDaily to over 0.94 on Beauty and Sports.
This confirms that temporal evaluation is much more challenging than closed-world leave-one-out evaluation, where many test targets may already be covered by the training item universe.
The performance gap between seen and unseen targets is substantial.
On seen targets, both TIGER and SASRec retain non-trivial performance, with dataset-dependent relative strengths: SASRec performs better on the Amazon datasets, while TIGER is stronger on Sephora, WikiLife, WeiboDaily, and WeiboTech.
However, both models largely collapse on unseen targets.
TIGER's unseen Recall@20 is nearly zero on most datasets.
SASRec shows a similar pattern, with unseen Recall@20 below 0.001 on Beauty, Sports, Toys, WeiboDaily, and WeiboTech.

These results show that temporal cold start exposes a reachability bottleneck hidden by seen-item evaluation.
Strong seen-target performance does not imply the ability to recommend items emerging after training.
For TIGER, the near-zero unseen performance suggests that SID-based generation is severely constrained outside the training item universe.
This motivates a finer question: are these failures caused by item-level novelty alone, or by limitations in SID token support and autoregressive path reachability?
We therefore proceed to a token-level diagnosis.

\subsection{Token-Level Coldness Taxonomy}

A cold item is not uniformly cold from the perspective of SID generation.
Let $\mathbf{c}(i)=(c_1(i),\ldots,c_L(i))$ be the SID of a cold item $i$.
We define the following token-level coldness categories.

\paragraph{All-token-seen.}
A cold item is all-token-seen if every atomic token in its SID appeared in training:
\begin{equation}
    i \in \mathcal{C}^{w}_{\mathrm{all\text{-}seen}}
    \Longleftrightarrow
    i \in \mathcal{I}^{w}_{\mathrm{cold}}
    \land
    \forall \ell,\;
    c_\ell(i) \in \mathcal{V}^{w}_{\mathrm{train}}.
\end{equation}
These items test whether the model can compose new SID combinations from seen atomic tokens.

\paragraph{Any-token-unseen.}
A cold item is any-token-unseen if at least one atomic token in its SID was absent from training:
\begin{equation}
    i \in \mathcal{C}^{w}_{\mathrm{any\text{-}unseen}}
    \Longleftrightarrow
    i \in \mathcal{I}^{w}_{\mathrm{cold}}
    \land
    \exists \ell,\;
    c_\ell(i) \notin \mathcal{V}^{w}_{\mathrm{train}}.
\end{equation}
These items require the model to handle token support that is unavailable or weakly parameterized during training.

\paragraph{Prefix-seen.}
A cold item is prefix-seen at length $\ell$ if its prefix has appeared in training:
\begin{equation}
    i \in \mathcal{C}^{w}_{\mathrm{prefix\text{-}seen}(\ell)}
    \Longleftrightarrow
    i \in \mathcal{I}^{w}_{\mathrm{cold}}
    \land
    (c_1(i),\ldots,c_\ell(i)) \in \mathcal{P}^{w}_{\ell}.
\end{equation}
Prefix-seen cold items test whether the model can complete a known coarse semantic path into a new leaf item.

\begin{table}[t]
\centering
% \vskip -0.05in
\caption{Representative token-level coldness taxonomy for unseen target items. We report the ratio of each coldness category and TIGER's Recall@20 and NDCG@20. Full results across all datasets are provided in \cref{tab:coldness_breakdown_full}.}
\vskip -0.15in
\label{tab:coldness_breakdown}
\resizebox{0.9\columnwidth}{!}{
\begin{tabular}{l l c cc}
\toprule
\textbf{Dataset}
& \textbf{Coldness category}
& \textbf{Ratio}
& \textbf{R@20}
& \textbf{N@20} \\
\midrule

\multirow{5}{*}{Sports}
& all-token-seen            & 0.405& 0.00006 & 0.00003 \\
& any-token-unseen          & 0.595& \z & \z \\
& prefix-seen-$c_1$         & 0.723& 0.00004 & 0.00002 \\
& prefix-seen-$c_1,c_2$     & 0.208& 0.00008 & 0.00004 \\
& prefix-seen-$c_1,c_2,c_3$ & 0.002& 0.01007 & 0.00520 \\
\midrule

\multirow{5}{*}{WeiboDaily}
& all-token-seen            & 1.000& 0.00092 & 0.00026 \\
& any-token-unseen          & 0.000& --      & --      \\
& prefix-seen-$c_1$         & 1.000& 0.00092 & 0.00026 \\
& prefix-seen-$c_1,c_2$     & 0.516& 0.00138 & 0.00039 \\
& prefix-seen-$c_1,c_2,c_3$ & 0.010& 0.01386 & 0.00379 \\

\bottomrule
\end{tabular}}
\vskip -0.2in
\end{table}

\begin{table*}[t]
\centering
\vskip -0.1in
\caption{Oracle-prefix probing for semantic ID generation.
We report both next-item semantic ID generation accuracy and next-token semantic ID generation accuracy.
``Free'' denotes standard autoregressive decoding, while ``Oracle-prefix'' decoding forces the corresponding ground-truth prefix before generating the remaining tokens.
$\Delta$ denotes the absolute improvement over free decoding. Darker red indicates larger improvement. Full results across all datasets are provided in \cref{tab:oracle_prefix_combined_full}.}
\label{tab:oracle_prefix_combined}
\vskip -0.1in
\resizebox{\textwidth}{!}{
\begin{tabular}{l l c c cc cc c ccc ccc}
\toprule
\multirow{3}{*}{\textbf{Dataset}}
& \multirow{3}{*}{\textbf{Group}}
& \multirow{3}{*}{\textbf{Ratio}}
& \multicolumn{5}{c}{\textbf{Next-item SID generation}}
& \multicolumn{7}{c}{\textbf{Next-token SID generation}} \\
\cmidrule(lr){4-8}\cmidrule(lr){9-15}
&
&
& \multirow{2}{*}{\textbf{Free}}
& \multicolumn{2}{c}{\textbf{Oracle $c_1$}}
& \multicolumn{2}{c}{\textbf{Oracle $c_1,c_2$}}
& \multicolumn{1}{c}{\textbf{$c_1$ Acc.}}
& \multicolumn{3}{c}{\textbf{$c_2$ Acc.}}
& \multicolumn{3}{c}{\textbf{$c_3$ Acc.}} \\
\cmidrule(lr){5-6}\cmidrule(lr){7-8}
\cmidrule(lr){9-9}\cmidrule(lr){10-12}\cmidrule(lr){13-15}
&
&
&
& Acc. & $\Delta$
& Acc. & $\Delta$
& Free
& Free & Oracle $c_1$ & $\Delta$
& Free & Oracle $c_1,c_2$ & $\Delta$ \\
\midrule

\multirow{2}{*}{Sports}
& Seen   & 0.053& 0.00063 & 0.00630 & \dlight{0.00567} & 0.02815 & \dmid{0.02752} & 0.18009 & 0.04449 & 0.07936 & \dmid{0.03487} & 0.00284 & 0.02910 & \dmid{0.02626} \\
& Unseen & 0.947& 0.00001 & 0.00006 & \dlight{0.00005} & 0.00448 & \dlight{0.00447} & 0.11983 & 0.00494 & 0.01091 & \dlight{0.00597} & 0.00138 & 0.00530 & \dlight{0.00392} \\
\midrule

% \multirow{2}{*}{Toys}
% & Seen   & 0.09000 & 0.00057 & 0.00564 & \dlight{0.00506} & 0.09664 & \dmid{0.09607} & 0.25815 & 0.01563 & 0.04140 & \dmid{0.02577} & 0.00514 & 0.09814 & \dmid{0.09300} \\
% & Unseen & 0.91000 & 0.00000 & 0.00000 & \dlight{0.00000} & 0.00272 & \dlight{0.00272} & 0.19895 & 0.00587 & 0.01855 & \dmid{0.01268} & 0.00391 & 0.00334 & -0.00057 \\
% \midrule

\multirow{2}{*}{WeiboDaily}
& Seen   & 0.747& 0.12319 & 0.47262 & \ddeep{0.34943} & 0.75122 & \ddeep{0.62804} & 0.18646 & 0.13814 & 0.49332 & \ddeep{0.35518} & 0.12957 & 0.75163 & \ddeep{0.62206} \\
& Unseen & 0.253& 0.00003 & 0.00049 & \dlight{0.00046} & 0.01493 & \dmid{0.01490} & 0.06264 & 0.01797 & 0.02774 & \dlight{0.00977} & 0.01480 & 0.01972 & \dlight{0.00492} \\
\midrule

\multirow{2}{*}{WeiboTech}
& Seen   & 0.615& 0.00632 & 0.07472 & \dmid{0.06841} & 0.39882 & \ddeep{0.39251} & 0.08608 & 0.01498 & 0.10570 & \dmid{0.09071} & 0.01139 & 0.39933 & \ddeep{0.38794} \\
& Unseen & 0.385& \z & 0.00004 & \dlight{0.00004} & 0.01173 & \dmid{0.01173} & 0.06236 & 0.00881 & 0.01749 & \dlight{0.00868} & 0.00647 & 0.01666 & \dmid{0.01020} \\

\bottomrule
\end{tabular}}
\vskip -0.1in
\end{table*}

\textbf{Results.}
\cref{tab:coldness_breakdown} reports the token-level coldness taxonomy for unseen target items.
The results show that temporal cold start is not a binary condition.
Cold-item reachability depends on multiple levels of support: whether the item ID is unseen, whether its atomic SID tokens are observed, and whether its SID prefixes follow paths supported by training data.

\textbf{Atomic token coverage is necessary but not sufficient.}
If SID generation were fully compositional, all-token-seen cold items should be relatively easy to reach.
However, TIGER still achieves almost zero performance in this regime on several datasets.
For example, in Sports, 40.5\% of cold items are all-token-seen, yet Recall@20 and NDCG@20 remain almost 0.
WeiboDaily shows a similar pattern: all cold items are all-token-seen, but TIGER still performs poorly.
This indicates that observing individual SID tokens during training does not guarantee that the decoder can compose them into a valid cold-item path.
A possible explanation is the path-dependent nature of RQ-VAE-style SID construction: later codes encode residuals left by earlier codes, so a SID forms an ordered residual path rather than a set of independently reusable tokens.
Thus, a new combination of seen tokens may still correspond to a weakly supported or unsupported path, making deeper prefix support more predictive than atomic token coverage alone.

\textbf{Unseen atomic tokens form a harder barrier.}
For any-token-unseen cold items, performance is consistently zero across datasets.
This is expected, since the decoder receives little or no supervision for generating tokens absent from the training targets.
Such items therefore lie outside the reliable generation space of the current SID-based model.

\textbf{Successful hits concentrate in prefix-supported regions.}
The prefix-based categories show that TIGER succeeds mainly when a cold item shares sufficiently deep SID prefixes with training items.
For example, in WeiboDaily, all cold items are all-token-seen, but Recall@20 increases from 0.00092 for prefix-seen-$c_1$ to 0.01386 for prefix-seen-$c_1,c_2,c_3$.
Sports shows a similar pattern, where the prefix-seen-$c_1,c_2,c_3$ subset achieves much higher Recall@20 than shallower prefix groups.
Although these deep-prefix subsets are small, they suggest that cold-item hits are driven more by path-level support than by free recombination of atomic tokens.

\textbf{SID-based cold-item reachability is narrowly bounded.}
Overall, TIGER can occasionally reach cold items whose tokens and deeper prefixes are sufficiently supported by training data.
However, it struggles with both unseen atomic tokens and new SID paths formed from seen tokens.
This supports the view that SID generation is constrained traversal over a learned token-path space, rather than open-ended semantic item generation.
This pattern motivates our later variants, which aim to reduce SID path dependency, relax exact decoding into scoring, and enrich generation with dynamic contextual signals.

\subsection{Oracle-Prefix Probing}

The previous analysis shows which cold items are easier or harder.
We next ask where the generation failure occurs.
Given a target SID $\mathbf{c}(i)=(c_1(i),\ldots,c_L(i))$, we compare free decoding with oracle-prefix decoding.
In free decoding, the model generates all tokens:
\begin{equation}
    \hat{\mathbf{c}}
    =
    \arg\max_{\mathbf{c}}
    \prod_{\ell=1}^{L}
    p_{\theta}(c_\ell \mid \mathcal{H}_u, c_{<\ell}).
\end{equation}

In oracle-$m$ decoding, the first $m$ target tokens are forced to be correct, and the model only generates the remaining suffix:
\begin{equation}
    \hat{\mathbf{c}}_{>m}
    =
    \arg\max_{\mathbf{c}_{>m}}
    \prod_{\ell=m+1}^{L}
    p_{\theta}(c_\ell \mid \mathcal{H}_u, c_{1:m}^{\star}(i), c_{m+1:\ell-1}),
\end{equation}

where $c_{1:m}^{\star}(i)$ is the ground-truth prefix.

\textbf{Results.}
\cref{tab:oracle_prefix_combined} reports oracle-prefix probing from both next-item SID generation and next-token generation perspectives.
Free decoding denotes standard autoregressive generation, while oracle-prefix decoding forces the first one or two ground-truth SID tokens before generating the suffix.
This setup separates errors in coarse bucket selection from failures in fine-grained SID path completion.

\textbf{Oracle prefixes strongly improve seen-item generation.}
For seen targets, forcing the correct prefix consistently improves exact SID accuracy, especially when the first two tokens are given.
For example, on WeiboDaily, accuracy increases from 0.12319 under free decoding to 0.47262 with oracle-$c_1$, and further to 0.75122 with oracle-$c_1,c_2$.
Oracle-$c_1,c_2$ also leads to significant increases in seen-item accuracy on other datasets.
This indicates that, within the training-supported item space, TIGER can often complete the remaining SID path once routed to a sufficiently specific prefix.

\textbf{Item-level unseen does not imply token-level unseen.}
For unseen targets, first-token accuracy can still be non-trivial, reaching 0.29701 on Sephora, 0.27493 on WikiLife, and 0.11983 on Sports.
This is because the seen/unseen split is defined at the item level: an unseen item may still share coarse SID tokens with training items.
Thus, the main difficulty is not always coarse bucket prediction; the model may identify the correct semantic region but still fail to generate the exact full SID.
The few slightly negative $\Delta$ values in next-token probing are also consistent with this view: oracle forcing changes the conditioning prefix, but the remaining suffix may still be weakly supported, so prefix correction does not guarantee monotonic improvement for every later token.

\textbf{The main bottleneck is fine-grained path completion.}
Even with oracle-$c_1,c_2$, unseen next-item SID accuracy remains very low: 0.00448 on Sports, 0.01493 on WeiboDaily, and 0.01173 on WeiboTech.
These values are far below the corresponding seen-item results.
Thus, correcting early bucket selection can recover many seen targets, but it is insufficient for item-level cold targets whose suffixes often involve weakly supported atomic tokens, unsupported SID paths, or item semantic identities absent from training.

\textbf{Cold-item reachability is path-bounded.}
Oracle-prefix probing shows that SID-based GR does not fail merely because it chooses the wrong first token.
For seen items, oracle prefixes reveal strong completion ability over supported SID paths.
For item-level unseen items, coarse semantic tokens may still be predictable, but exact reachability remains limited by path-level support.
This further supports our view that SID generation enables token-supported recombination, while cold-item reachability is bounded by the learned SID path structure rather than atomic token availability alone.

\section{From Generation to Hierarchical Semantic Bucketing}
\label{semantic_bucketing}
The token-level diagnosis suggests that SID-based GR does not behave like unconstrained natural language generation.
Instead, its output space forms a structured semantic code tree, where each level corresponds to a semantic granularity.
We therefore reinterpret autoregressive SID decoding as hierarchical semantic bucket classification:
the first token selects a coarse semantic bucket, and subsequent tokens recursively refine it into finer sub-buckets until a complete SID is reached.

Let the SID of item $i$ be denoted as $\mathbf{c}(i)=(c_1(i),\ldots,c_L(i))$.
For TIGER, the token distribution at depth $\ell$ given the prefix $\mathbf{a}_{<\ell} = (c_1, \ldots, c_{\ell-1})$ is defined as:
\begin{equation}
    P_{\mathrm{TIGER}}^{(\ell)}(b \mid u,\mathbf{a}_{<\ell})
    =
    p_{\theta}(c_\ell=b \mid \mathcal{H}_u,\mathbf{a}_{<\ell}),
\end{equation}
where $\mathbf{a}_{<1}$ is empty for the first SID position, and $b$ represents a candidate sub-bucket index at depth $\ell$.
This distribution reflects TIGER's belief over the next semantic sub-bucket.

To compare this generative decision process with a discriminative recommender, we aggregate SASRec's item-level scores into the same SID bucket space.
Let $\pi_\phi(i\mid u)$ denote the softmax-normalized SASRec probability over the candidate item set $\mathcal{I}_{\mathrm{cand}}$.
Given a prefix $\mathbf{a}_{<\ell}$, we define the set of prefix-compatible items as:
\begin{equation}
    \mathcal{I}(\mathbf{a}_{<\ell})
    =
    \left\{
    i\in\mathcal{I}_{\mathrm{cand}}
    \mid
    (c_1(i),\ldots,c_{\ell-1}(i))=\mathbf{a}_{<\ell}
    \right\}.
\end{equation}
The corresponding bucket distribution for SASRec is then computed by re-normalizing the probabilities of compatible items:
\begin{equation}
    P_{\mathrm{SASRec}}^{(\ell)}(b \mid u,\mathbf{a}_{<\ell})
    =
    \frac{
    \sum_{i\in \mathcal{I}(\mathbf{a}_{<\ell}),\, c_\ell(i)=b}
    \pi_\phi(i \mid u)
    }{
    \sum_{i\in \mathcal{I}(\mathbf{a}_{<\ell})}
    \pi_\phi(i \mid u)
    }.
\end{equation}
We compare $P_{\mathrm{TIGER}}^{(\ell)}$ and $P_{\mathrm{SASRec}}^{(\ell)}$ using Jensen--Shannon divergence and Pearson correlation.
Lower JS divergence and higher correlation indicate stronger distributional alignment between generative token prediction and discriminative bucket scoring.

\begin{figure}[t]
\centering
% \vskip -0.1in
\includegraphics[width=0.75\columnwidth]{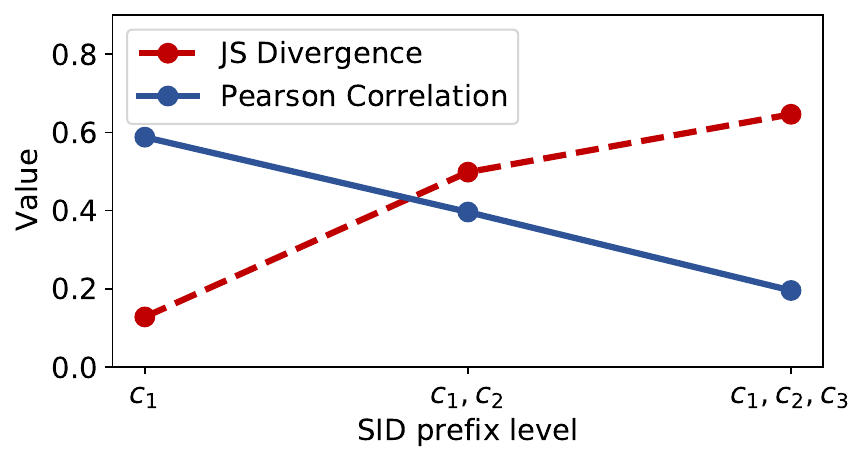}
\vskip -0.1in
\caption{Distributional alignment between TIGER prefix generation and SASRec bucket-aggregated scoring on Beauty. Full numerical results are shown in \cref{tab:bucket_alignment}.}
\vskip -0.1in
\label{fig:beauty_js_pearson_combined}
\end{figure}

% \noindent
\textbf{Results.}
As shown in \cref{fig:beauty_js_pearson_combined,tab:bucket_alignment}, the alignment exhibits a consistent coarse-to-fine pattern.
At shallow SID positions, TIGER's token distribution is close to SASRec's bucket-aggregated distribution, suggesting that early tokens mainly capture coarse semantic preferences.
As the hierarchy deepens, JS divergence increases and Pearson correlation decreases, indicating that later tokens perform increasingly fine-grained semantic refinement.
This supports the view that SID-based generation is better understood as hierarchical semantic bucketing rather than open-ended item generation.
It also explains the cold-start boundary: a cold item is reachable only when its SID lies on a sufficiently supported semantic path.
Thus, the open-world reach of SID-based GR is constrained by the learned bucket hierarchy.

\section{Reachability-Oriented TIGER Variants}\label{variant_demo}
% \begin{table*}[t]
% \centering
% \vskip -0.1in
% \caption{Seen and unseen recommendation performance of TIGER-based diagnostic variants under temporal cold-start evaluation. Full results are shown in \cref{tab:variant_seen_unseen_full}.}
% \vskip -0.1in
% \label{tab:variant_seen_unseen}
% \resizebox{\textwidth}{!}{
% \begin{tabular}{l c cc cc cc cc}
% \toprule
% \multirow{2}{*}{\textbf{Dataset}}
% & \multirow{2}{*}{\textbf{Cold ratio}}
% & \multicolumn{2}{c}{\textbf{TIGER}}
% & \multicolumn{2}{c}{\textbf{TIGER-SID}}
% & \multicolumn{2}{c}{\textbf{TIGER-Scorer}}
% & \multicolumn{2}{c}{\textbf{TIGER-Edge}} \\
% \cmidrule(lr){3-4}\cmidrule(lr){5-6}\cmidrule(lr){7-8}\cmidrule(lr){9-10}
% &
% & Seen H@20 & Unseen H@20
% & Seen H@20 & Unseen H@20
% & Seen H@20 & Unseen H@20
% & Seen H@20 & Unseen H@20 \\
% \midrule
% Sports      & 0.947& 0.01406 & 0.00020 & 0.01582 & 0.00109 & 0.04120 & 0.01392 & 0.03886 & 0.00075 \\
% \midrule
% WeiboDaily  & 0.253& 0.75894 & 0.00313 & 0.77940 & 0.00747 & 0.32633 & 0.00750 & 0.81308 & 0.01132 \\
% WeiboTech   & 0.385& 0.41039 & \z & 0.49199 & 0.01232 & 0.16490 & 0.02929 & 0.47153 & 0.01683 \\
% \bottomrule
% \end{tabular}}
% \end{table*}

\begin{table*}[t]
\centering
\vskip -0.1in
\caption{Seen and unseen recommendation performance of TIGER-based diagnostic variants under temporal cold-start evaluation. Full results are shown in \cref{tab:variant_seen_unseen_full}. Arrows indicate relative change compared to TIGER (\textcolor{red}{$\uparrow$},\textcolor{blue}{$\downarrow$}: >100\%, \textcolor{red!40}{$\uparrow$},\textcolor{blue!40}{$\downarrow$}: $\le$100\%).}
\vskip -0.1in
\label{tab:variant_seen_unseen}
\resizebox{\textwidth}{!}{
\begin{tabular}{l c cc cc cc cc}
\toprule
\multirow{2}{*}{\textbf{Dataset}}
& \multirow{2}{*}{\textbf{Cold ratio}}
& \multicolumn{2}{c}{\textbf{TIGER}}
& \multicolumn{2}{c}{\textbf{TIGER-SID}}
& \multicolumn{2}{c}{\textbf{TIGER-Scorer}}
& \multicolumn{2}{c}{\textbf{TIGER-Edge}} \\
\cmidrule(lr){3-4}\cmidrule(lr){5-6}\cmidrule(lr){7-8}\cmidrule(lr){9-10}
&
& Seen H@20 & Unseen H@20
& Seen H@20 & Unseen H@20
& Seen H@20 & Unseen H@20
& Seen H@20 & Unseen H@20 \\
\midrule
Beauty      & 0.948 & 0.01334 & 0.00023 
& 0.00655\,\textcolor{blue!40}{$\downarrow$} & 0.00054\,\textcolor{red!40}{$\uparrow$} 
& \textbf{0.02163}\,\textcolor{red!40}{$\uparrow$} & \textbf{0.00831}\,\textcolor{red}{$\uparrow$} 
& {0.01925}\,\textcolor{red!40}{$\uparrow$} & 0.00042\,\textcolor{red!40}{$\uparrow$} \\

Sports      & 0.947& 0.01406 & 0.00020 
& 0.01582\,\textcolor{red!40}{$\uparrow$} & 0.00109\,\textcolor{red}{$\uparrow$} 
& \textbf{0.04120}\,\textcolor{red}{$\uparrow$} & \textbf{0.01392}\,\textcolor{red}{$\uparrow$} 
& 0.03886\,\textcolor{red}{$\uparrow$} & 0.00075\,\textcolor{red}{$\uparrow$} \\

\midrule
WeiboDaily  & 0.253& 0.75894 & 0.00313 
& 0.77940\,\textcolor{red!40}{$\uparrow$} & 0.00747\,\textcolor{red}{$\uparrow$} 
& 0.32633\,\textcolor{blue!40}{$\downarrow$} & 0.00750\,\textcolor{red}{$\uparrow$} 
& \textbf{0.81308}\,\textcolor{red!40}{$\uparrow$} & \textbf{0.01132}\,\textcolor{red}{$\uparrow$} \\

WeiboTech   & 0.385& 0.41039 & \z 
& \textbf{0.49199}\,\textcolor{red!40}{$\uparrow$} & 0.01232\,\textcolor{red}{$\uparrow$} 
& 0.16490\,\textcolor{blue!40}{$\downarrow$} & \textbf{0.02929}\,\textcolor{red}{$\uparrow$} 
& {0.47153}\,\textcolor{red!40}{$\uparrow$} & 0.01683\,\textcolor{red}{$\uparrow$} \\

\bottomrule
\end{tabular}}
\end{table*}

% \begin{table*}[t]
% \centering
% \vskip -0.1in
% \caption{Token-level coldness taxonomy for unseen target items on each variant. Full results are shown in \cref{tab:variant_token_support_full}.}
% \vskip -0.1in
% \label{tab:variant_token_support}
% \resizebox{0.9\textwidth}{!}{
% \begin{tabular}{l l cc cc cc cc}
% \toprule
% \multirow{2}{*}{\textbf{Dataset}}
% & \multirow{2}{*}{\textbf{Coldness category}}
% & \multicolumn{2}{c}{\textbf{TIGER}}
% & \multicolumn{2}{c}{\textbf{TIGER-SID}}
% & \multicolumn{2}{c}{\textbf{TIGER-Scorer}}
% & \multicolumn{2}{c}{\textbf{TIGER-Edge}} \\
% \cmidrule(lr){3-4}\cmidrule(lr){5-6}\cmidrule(lr){7-8}\cmidrule(lr){9-10}
% &
% & Recall@5 & Recall@20
% & Recall@5 & Recall@20
% & Recall@5 & Recall@20
% & Recall@5 & Recall@20 \\
% \midrule

% \multirow{5}{*}{Sports}
% & all-token-seen                & 0.00001 & 0.00006 & \z & 0.00019 & 0.00014 & 0.00210 & 0.00004 & 0.00024 \\
% & any-token-unseen              & \z & \z & \z & \z & 0.00254 & 0.00274 & \z & \z \\
% & prefix-seen-$c_1$             & 0.00001 & 0.00004 & \z & 0.00017 & 0.01343 & 0.01441 & 0.00003 & 0.00016 \\
% & prefix-seen-$c_1,c_2$         & 0.00002 & 0.00008 & \z & 0.00035 & 0.16681 & 0.16971 & 0.00006 & 0.00036 \\

% \bottomrule
% \end{tabular}}
% \vskip -0.1in
% \end{table*}

\begin{table*}[t]
\centering
\vskip -0.1in
\caption{Token-level coldness taxonomy for unseen target items on each variant. Full results are shown in \cref{tab:variant_token_support_full}. }
\vskip -0.1in
\label{tab:variant_token_support}
\resizebox{0.9\textwidth}{!}{
\begin{tabular}{l l cc cc cc cc}
\toprule
\multirow{2}{*}{\textbf{Dataset}}
& \multirow{2}{*}{\textbf{Coldness category}}
& \multicolumn{2}{c}{\textbf{TIGER}}
& \multicolumn{2}{c}{\textbf{TIGER-SID}}
& \multicolumn{2}{c}{\textbf{TIGER-Scorer}}
& \multicolumn{2}{c}{\textbf{TIGER-Edge}} \\
\cmidrule(lr){3-4}\cmidrule(lr){5-6}\cmidrule(lr){7-8}\cmidrule(lr){9-10}
&
& Recall@5 & Recall@20
& Recall@5 & Recall@20
& Recall@5 & Recall@20
& Recall@5 & Recall@20 \\
\midrule

\multirow{5}{*}{Sports}
& all-token-seen                
& 0.00001 & 0.00006 
& \z\,\textcolor{blue!40}{$\downarrow$} & 0.00019\,\textcolor{red!40}{$\uparrow$} 
& \textbf{0.00014}\,\textcolor{red}{$\uparrow$} & \textbf{0.00210}\,\textcolor{red}{$\uparrow$} 
& 0.00004\,\textcolor{red!40}{$\uparrow$} & 0.00024\,\textcolor{red!40}{$\uparrow$} \\

& any-token-unseen              
& \z & \z 
& \z & \z 
& \textbf{0.00254}\,\textcolor{red}{$\uparrow$} & \textbf{0.00274}\,\textcolor{red}{$\uparrow$} 
& \z & \z \\

& prefix-seen-$c_1$             
& 0.00001 & 0.00004 
& \z\,\textcolor{blue!40}{$\downarrow$} & 0.00017\,\textcolor{red!40}{$\uparrow$} 
& \textbf{0.01343}\,\textcolor{red}{$\uparrow$} & \textbf{0.01441}\,\textcolor{red}{$\uparrow$} 
& 0.00003\,\textcolor{red!40}{$\uparrow$} & 0.00016\,\textcolor{red!40}{$\uparrow$} \\

& prefix-seen-$c_1,c_2$         
& 0.00002 & 0.00008 
& \z\,\textcolor{blue!40}{$\downarrow$} & 0.00035\,\textcolor{red!40}{$\uparrow$} 
& \textbf{0.16681}\,\textcolor{red}{$\uparrow$} & \textbf{0.16971}\,\textcolor{red}{$\uparrow$} 
& 0.00006\,\textcolor{red!40}{$\uparrow$} & 0.00036\,\textcolor{red!40}{$\uparrow$} \\

\bottomrule
\end{tabular}}
\vskip -0.1in
\end{table*}

The previous sections show that the cold-start weakness of SID-based generation is not only an accuracy problem, but also a reachability problem.
A cold item can be recommended only if its SID lies in a token-path region that the generator can express and decode.
To connect this diagnosis with concrete design routes, we study three controlled TIGER-based variants.
They are not proposed as standalone recommender architectures; instead, they keep the TIGER backbone or objective as unchanged as possible and modify one component at a time.

\subsection{Overview of Diagnostic Variants}

Let $\mathcal{H}_u^{<t}$ denote the historical interactions of user $u$ before time $t$.
Standard TIGER predicts a target item $i$ by autoregressively generating its SID $\mathbf{c}(i)=[c_1(i),\ldots,c_L(i)]$.
% \begin{equation}
% p_{\theta}\!\left(\mathbf{c}(i)\mid \mathcal{H}_u^{<t}\right)
% =
% \prod_{\ell=1}^{L}
% p_{\theta}\!\left(
% c_{\ell}(i)
% \mid
% \mathcal{H}_u^{<t}, c_{<\ell}(i)
% \right).
% \end{equation}
This formulation couples recommendation success with two requirements: learning useful user preferences from interaction history, and reaching the target item through a supported autoregressive SID path.
Our temporal cold-start results suggest that the latter is often the limiting factor.
Even when coarse semantic preference is captured, the decoder may still fail to generate the exact SID path of an emerging item.

We therefore introduce three diagnostic variants that target different parts of this reachability bottleneck.
\textbf{TIGER-SID} changes the SID tokenizer while keeping autoregressive generation unchanged, testing whether a less path-dependent and more factorized SID space makes cold items easier to express through seen-token recombination.
\textbf{TIGER-Scorer} keeps the TIGER encoder but replaces exact SID generation with candidate scoring, testing whether useful preference representations are already learned even when exact SID decoding fails.
\textbf{TIGER-Edge} keeps the generative SID objective but augments decoding with dynamic edge or contextual information through cross attention, testing whether relational evidence can guide the generator toward better semantic regions.
Together, these variants separate three possible failure sources: SID-space expressiveness, autoregressive decoding rigidity, and deficiency in dynamic context.
Detailed architectures are provided in \cref{app:variant_details}.

\subsection{Evaluation Protocol and Results}

We evaluate all variants using the same absolute-time sliding-window protocol described previously.
For each test window, target items are divided into {seen} and {unseen} groups based on the presence of their item IDs in the training set.
\cref{tab:variant_seen_unseen} reports averaged seen/unseen Hit@20, and \cref{tab:variant_token_support} further breaks down cold-item performance by our token-level coldness taxonomy.
This allows us to examine not only whether a variant improves temporal cold-start performance, but also which part of the SID reachability boundary it expands.

\textbf{Results.} \cref{tab:variant_seen_unseen} shows that the three variants expand different parts of the boundary.
\textbf{TIGER-SID} brings clear gains on WeiboDaily and WeiboTech, showing that a more factorized SID space can make several future items easier to express, although SID redesign alone cannot thoroughly remove the decoding constraint.
\textbf{TIGER-Scorer} greatly improves performance on unseen items across several datasets. This indicates that while the TIGER encoder effectively captures user preferences, the autoregressive decoder inherently limits the reachability of cold items. Notably, the scoring mechanism incurs a performance trade-off on seen items, likely due to the extensive search space of the candidate set.
\textbf{TIGER-Edge} proves most effective on dynamic datasets characterized by rich contextual signals, demonstrating that integrating edge context can effectively guide the decoder toward more relevant semantic regions.

The results in \cref{tab:variant_token_support} further corroborate these findings.
Direct generation variants mainly help in token-supported or prefix-supported regimes, while any-token-unseen cases remain challenging.
Conversely, TIGER-Scorer mitigates exact-decoding failures by ranking candidates even in {any-token-unseen} scenarios. Meanwhile, TIGER-Edge shows improvements in {deeper prefix-supported} cases where relational context provides informative guidance.
Overall, these variants do not entirely eliminate the cold-item reachability boundary, but reveal three complementary ways to expand it: more compositional SID codes, scoring-based prediction, and contextual enrichment.
\section{Design Implications}\label{design_implication}

Our diagnosis and TIGER-based variants suggest a reachability-oriented view of temporal cold start.
The central issue is not only low accuracy on unseen items, but that future items may fall outside the model's reachable token-path space.
This boundary is mainly shaped by three factors: SID representation, prediction interface, and inference-time context.

\textbf{SID-space design.}
TIGER-SID shows that item encoding directly affects cold-item reachability.
Residual-quantized SIDs are compact, but their ordered paths can make future items hard to express even when individual tokens have been observed.
More factorized designs, such as product quantization and order-insensitive codebooks \cite{ge2013optimized,hou2025generating}, may reduce path dependency and make more future items token-supported or prefix-supported.
Thus, SID design should be evaluated not only by reconstruction or ranking accuracy, but also by its ability to expand future-item support.

\textbf{Prediction interface.}
TIGER-Scorer shows that failure can also arise from exact SID decoding.
An autoregressive generator must produce every target token correctly, so a single token error can prevent a useful user representation from reaching the correct item.
A scorer interface relaxes this constraint by scoring candidates with learned user representations.
This trades strict generation for broader reachability, making performance depend more on candidate coverage and quality.

\textbf{Dynamic context.}
TIGER-Edge suggests that future items should not be treated as static identifiers.
New items often arrive with text, metadata, temporal signals, or relational edges.
Using such evidence allows SID prediction to consider more contextual information.
This motivates the development of context-aware recommendation systems to evaluate items whose SID paths are weakly supported or entirely unsupported.

\subsection{A Unified View}

These routes are complementary.
SID-space design concerns whether future items are expressible.
Generator-versus-scorer design concerns whether the model must generate the exact SID path or can rank candidates through a softer interface.
Dynamic-context design concerns whether the system can exploit evidence beyond static item tokens as the catalog evolves.
Together, they define a reachability-oriented roadmap:
\begin{equation}
    \underbrace{\mathrm{SID}(i)}_{\text{expressibility}}
    +
    \underbrace{\mathrm{Predictor}(u,i)}_{\text{generation vs. scoring}}
    +
    \underbrace{\mathrm{Context}(u,i,t)}_{\text{dynamic evidence}}
    \Longrightarrow
    \mathrm{Reachability}(u,i,t).
\end{equation}

This view clarifies the role of our variants: they are controlled probes of the factors determining whether a future cold item can be reached.
The goal is therefore not only to report a cold-start accuracy drop, but to identify where the reachability boundary comes from and how future systems may expand it.
\section{Conclusion}\label{conclusion}
This work revisits SID-based generative recommendation under temporal cold start. Using an absolute-time protocol, we show that closed-world SID recombination does not necessarily imply inductive reachability for newly emerging items. Token-level coldness analysis, seen/unseen-hit diagnosis, and oracle-prefix probing reveal that current SID-based GR can reach cold items mainly when their tokens and prefixes are supported by past interactions, but struggles with unseen atomic tokens and unsupported paths. We further explain this boundary through hierarchical semantic bucketing: early SID tokens select coarse semantic regions, while later tokens refine item-specific paths. This view clarifies that SID generation provides semantic compression and compositional sharing, but remains bounded by learned token vocabularies and path support. Future work should therefore explore more inductive SID spaces, scoring-based verification, and dynamic item, edge, or textual context for evolving catalogs.
Overall, our findings call for generative recommenders that are better aligned with evolving item sets, where new items could be represented, verified, and incorporated beyond static SID sequences.

\appendix

\newpage
\twocolumn
% \balance
\bibliographystyle{ACM-Reference-Format}

\bibliography{reference}

\newpage

\section{TIGER-based Variant}\label{app:variant_details}

\subsection{TIGER-SID: Enlarging the Factorized SID Space}

TIGER-SID addresses the {reachability bottleneck} inherent in the SID representation design. 
In the original TIGER framework, item SIDs are typically generated via {Residual Quantization} (RQ). 
In this scheme, the code at a deeper position depends on the residual error left by preceding positions. 
Consequently, an item is not represented merely by a set of independent tokens, but by a strictly ordered residual path. 
For cold items, this path dependency imposes a significant constraint: even if every individual token has been observed during training, the specific ordered combination required to form the target path may remain difficult for the generator to decode correctly.

To mitigate this, TIGER-SID replaces the residual-style SID construction with a {factorized tokenizer} based on Optimized Product Quantization (OPQ)~\cite{ge2013optimized,hou2025generating}. 
While the model architecture and training objectives remain identical to the original TIGER (i.e., the T5-based generator autoregressively predicts the target SID from user history), the underlying code space topology is fundamentally altered. 
Instead of enforcing a dependency where each code explains a residual conditioned on previous codes, TIGER-SID employs OPQ to assign SID positions more independently. 
The item representation $\mathbf{z}_i$ is constructed by concatenating embeddings from $L$ independent subspaces:
\begin{equation}
    \mathbf{z}_i = [\mathbf{e}^{(1)}_{c_1(i)}; \mathbf{e}^{(2)}_{c_2(i)}; \ldots; \mathbf{e}^{(L)}_{c_L(i)}],
\end{equation}
where $c_l(i)$ denotes the code assigned to item $i$ at the $l$-th position, and $\mathbf{e}^{(l)}$ represents the corresponding codebook embedding. 

This design significantly expands the volume of theoretically valid item codes that can be synthesized by recombining tokens seen during training, as it eliminates the rigid constraints of ordered residual paths. 
Consequently, while the reachability of cold items still relies on the observability of constituent tokens, TIGER-SID alleviates the restrictive path dependency characteristic of RQ-VAE. 
Ultimately, TIGER-SID investigates whether recommendation performance can be improved simply by rendering the SID space more compositional and less prefix-dependent.

\subsection{TIGER-Scorer: Replacing Exact SID Decoding with Candidate Scoring}

TIGER-Scorer targets the {prediction interface} to mitigate the sensitivity of generative recommendation. 
The original generator is constrained to output the exact SID sequence of the target item. 
This imposes a rigid constraint where a single token error can divert the prediction to an irrelevant item or an invalid path, even if the encoder has successfully captured the user's coarse semantic preferences. 
To address this, TIGER-Scorer bypasses the autoregressive decoding process and reformulates TIGER as a {candidate scoring model}.

Concretely, TIGER-Scorer preserves the T5 encoder from the original architecture. 
The encoder maps the user history $\mathcal{H}_u^{<t}$ into a normalized user representation $\mathbf{u}$. 
Simultaneously, each candidate item $i \in \mathcal{C}_t$ is mapped into the same latent space by fusing its SID embedding and its content embedding to form a unified item representation $\mathbf{v}_i$. 
The recommendation score is then computed via a temperature-scaled dot product:
\begin{equation}
    s(u,i) = \frac{\mathbf{u}^{\top}\mathbf{v}_i}{\tau}, \quad \forall i \in \mathcal{C}_t,
\end{equation}
where $\tau$ is a temperature hyperparameter controlling the sharpness of the probability distribution. 
The model is optimized using cross-entropy loss over the candidate set and performs inference by ranking candidates according to $s(u,i)$.

Formally, this shifts the inference paradigm from exact SID decoding, denoted as $\mathcal{H}_u^{<t} \rightarrow \hat{\mathbf{c}} \rightarrow \hat{i}$, to direct candidate scoring: $(\mathcal{H}_u^{<t}, \mathcal{C}_t) \rightarrow \{s(u,i)\}_{i \in \mathcal{C}_t} \rightarrow \hat{i}$. 
Consequently, TIGER-Scorer effectively disentangles whether cold-start failures stem from inadequate user preference modeling or the rigid requirement of generating the exact SID sequence.

\subsection{TIGER-Edge: Injecting Dynamic Edge Context via Cross-Attention}

TIGER-Edge targets both the input signal and the model architecture.
In dynamic recommendation settings, cold items are often accompanied by rich contextual information, such as interaction text, edge attributes, temporal signals, relation types, or neighborhood evidence.
The standard SID generator primarily conditions on the historical item sequence, potentially overlooking such dynamic relational signals.
TIGER-Edge addresses this limitation by augmenting the generator with an external context stream, enabling the sequence representation to attend to this additional information.

Specifically, TIGER-Edge employs a GPT-style SID decoder and enriches it with an external context stream derived from historical edge or text representations.
Given the raw edge/context features $\mathbf{X}^{\mathrm{edge}}_u$ for user $u$, the model first projects them separately into the decoder's hidden space for keys and values:
\begin{equation}
    \mathbf{K}_u = \mathrm{Proj}_K(\mathbf{X}^{\mathrm{edge}}_u), \quad
    \mathbf{V}_u = \mathrm{Proj}_V(\mathbf{X}^{\mathrm{edge}}_u),
\end{equation}
where $\mathrm{Proj}_K$ and $\mathrm{Proj}_V$ are two independent learnable linear projections.
A gated cross-attention layer then allows the sequential hidden states to attend to this external context:
\begin{equation}
    \mathbf{A}_u = \mathrm{Attn}(\mathbf{H}_u, \mathbf{K}_u, \mathbf{V}_u),
\end{equation}
where $\mathbf{H}_u$ represents the decoder hidden states, and $\mathbf{K}_u, \mathbf{V}_u$ serve as the key and value contexts derived from edge embeddings.
The attended representation $\mathbf{A}_u$ is subsequently used for SID decoding.

Thus, TIGER-Edge preserves the original generative SID objective while providing the decoder with additional dynamic relational signals.
It investigates whether cold-start reachability can be improved by enriching the generator's context, rather than by modifying the SID tokenizer or replacing generation with candidate scoring.

\section{Dataset}\label{app:dataset}
\subsection{Dataset Details}
\label{app:dataset}

\begin{table*}[t]
\centering
\caption{Dataset statistics under the absolute-time sliding-window temporal cold-start protocol. 
Each dataset is divided into five chronological windows $W_1,\ldots,W_5$. 
We evaluate on Test Window $W_3$, $W_4$, and $W_5$, where each test window uses its preceding windows for training and validation.
Cold ratio denotes the cold-item proportion in the corresponding test split.}
\label{tab:sliding_window_dataset_statistics}
\resizebox{\textwidth}{!}{%
\begin{tabular}{l c c c | c c c| c c}
\toprule
\textbf{Dataset} 
& \textbf{Split}
& \textbf{Train window}
& \textbf{Test window}
& \textbf{Train sequences}
& \textbf{Val sequences}
& \textbf{Test sequences}
& \textbf{Test targets}
& \textbf{Cold ratio} \\
\midrule

\multicolumn{9}{l}{\textbf{Test Window $W_3$}} \\
\midrule
Beauty      & 1 & $W_1$       & $W_3$ & 12    & 5    & 56    & 254    & 0.984 \\
Sports      & 1 & $W_1$       & $W_3$ & 3     & 2    & 37    & 131    & 1.000 \\
Toys        & 1 & $W_1$       & $W_3$ & 101   & 57   & 324   & 900    & 0.903 \\
\midrule
Sephora     & 1 & $W_1$       & $W_3$ & 5,091 & 725  & 1,454 & 2,917  & 0.634 \\
WikiLife    & 1 & $W_1$       & $W_3$ & 509   & 36   & 113   & 286    & 0.650 \\
WeiboDaily  & 1 & $W_1$       & $W_3$ & 854   & 506  & 909   & 37,147 & 0.285 \\
WeiboTech   & 1 & $W_1$       & $W_3$ & 134   & 63   & 101   & 14,764 & 0.469 \\

\midrule
\multicolumn{9}{l}{\textbf{Test Window $W_4$}} \\
\midrule
Beauty      & 2 & $W_{1:2}$   & $W_4$ & 86     & 56    & 794   & 2,828  & 0.957 \\
Sports      & 2 & $W_{1:2}$   & $W_4$ & 69     & 37    & 641   & 2,379  & 0.960 \\
Toys        & 2 & $W_{1:2}$   & $W_4$ & 684    & 324   & 859   & 3,642  & 0.901 \\
\midrule
Sephora     & 2 & $W_{1:2}$   & $W_4$ & 12,523 & 1,454 & 5,337 & 14,274 & 0.823 \\
WikiLife    & 2 & $W_{1:2}$   & $W_4$ & 1,650  & 113   & 451   & 1,348  & 0.593 \\
WeiboDaily  & 2 & $W_{1:2}$   & $W_4$ & 1,597  & 909   & 1,233 & 59,954 & 0.247 \\
WeiboTech   & 2 & $W_{1:2}$   & $W_4$ & 216    & 101   & 212   & 21,740 & 0.409 \\

\midrule
\multicolumn{9}{l}{\textbf{Test Window $W_5$}} \\
\midrule
Beauty      & 3 & $W_{1:3}$   & $W_5$ & 1,043  & 794   & 4,434  & 36,100 & 0.904 \\
Sports      & 3 & $W_{1:3}$   & $W_5$ & 904    & 641   & 6,836  & 40,569 & 0.883 \\
Toys        & 3 & $W_{1:3}$   & $W_5$ & 1,538  & 859   & 3,959  & 31,265 & 0.925 \\
\midrule
Sephora     & 3 & $W_{1:3}$   & $W_5$ & 27,351 & 5,337 & 21,944 & 84,616 & 0.863 \\
WikiLife    & 3 & $W_{1:3}$   & $W_5$ & 5,152  & 451   & 4,339  & 10,088 & 0.463 \\
WeiboDaily  & 3 & $W_{1:3}$   & $W_5$ & 2,457  & 1,233 & 1,775  & 99,701 & 0.228 \\
WeiboTech   & 3 & $W_{1:3}$   & $W_5$ & 438    & 212   & 405    & 41,099 & 0.277 \\

\bottomrule
\end{tabular}%
}
\end{table*}

We conduct experiments on seven datasets, including three widely used Amazon review datasets and four recent Dynamic Text-attributed Graph (DyTAG) datasets. For all datasets, timestamped interactions are sorted chronologically to construct next-item prediction sequences.

\begin{itemize}
\item \textbf{Beauty, Sports, and Toys.}
These three datasets are adopted from the Amazon review collection~\cite{mcauley2015image}, and correspond to the Beauty, Sports and Outdoors, and Toys and Games product categories, respectively. Each dataset consists of timestamped user--product review interactions. We treat products as recommendable items and use user review histories to construct sequential recommendation instances. Product metadata and review information provide textual signals for semantic item representation.

\item \textbf{Sephora.}
Sephora is a DyTAG dataset from GDGB~\cite{peng2025gdgb}, built from beauty and skincare reviews on the Sephora platform. It is represented as a bipartite user--product graph, where users and products are textual nodes, and each edge corresponds to a timestamped rating and textual review. In our recommendation setting, Sephora products are treated as target items.

\item \textbf{WikiLife.}
WikiLife is a DyTAG dataset from GDGB~\cite{peng2025gdgb}, derived from Wikipedia life trajectories. It forms a bipartite person--location graph, where nodes correspond to persons and locations, and edges describe temporally grounded life activities of persons at specific locations. We treat locations as recommendable items and construct chronological person-to-location prediction sequences.

\item \textbf{WeiboDaily and WeiboTech.}
WeiboDaily and WeiboTech are DyTAG datasets from GDGB~\cite{peng2025gdgb}, constructed from timestamped user interactions on the Weibo social platform. Both datasets are non-bipartite user interaction graphs, where nodes are Weibo users and edges correspond to social interactions such as comments and reposts with associated textual content. WeiboDaily focuses more on daily-life topics, while WeiboTech is centered on technology-related topics. In our setting, destination users are treated as target items for next-item prediction.

\end{itemize}

We follow the same absolute-time sliding-window protocol for all datasets. Each dataset is partitioned into five chronological windows $W_1,\ldots,W_5$, and we evaluate on three test windows, namely $W_3$, $W_4$, and $W_5$. For test window $W_3$, the model is trained on $W_1$ and validated on $W_2$; for test window $W_4$, the model is trained on $W_1$--$W_2$ and validated on $W_3$; for test window $W_5$, the model is trained on $W_1$--$W_3$ and validated on $W_4$. This protocol preserves strict temporal order and ensures that test targets occur after the corresponding training period. Detailed statistics of the resulting splits, including the number of train/validation/test user sequences, test targets, and cold-item ratios, are reported in \cref{tab:sliding_window_dataset_statistics}. The dataset processing pipeline and source code are available at \url{https://github.com/Lucas-PJ/GRColdItemReachability}.

\section{Experimental Result}

\begin{itemize}[label=\textbullet, leftmargin=*, itemsep=0.5em, parsep=0.2em]
    \item \textbf{Temporal cold-start evaluation.} 
    Performance on seen and unseen targets, where ``Cold ratio'' denotes the average proportion of test targets whose item IDs are absent from the training period. H@K, R@K, and N@K denote Hit@K, Recall@K, and NDCG@K, respectively. Representative results are shown in \cref{tab:seen_unseen_main}. Detailed results for each window are presented in \cref{tab:tiger_seen_unseen_at5,tab:tiger_seen_unseen_at10,tab:tiger_seen_unseen_at20} for TIGER and \cref{tab:sasrec_seen_unseen_at5,tab:sasrec_seen_unseen_at10,tab:sasrec_seen_unseen_at20} for SASRec.

    \item \textbf{Token-level coldness taxonomy.} 
    Performance under the token-level coldness taxonomy for unseen target items, reporting the ratio of each coldness category along with TIGER's Recall@20 and NDCG@20. Representative results are shown in \cref{tab:coldness_breakdown}, while full results across all datasets are provided in \cref{tab:coldness_breakdown_full}.

    \item \textbf{Oracle-prefix probing for semantic ID generation.} We report both next-item and next-token semantic ID generation accuracy. ``Free'' denotes standard autoregressive decoding, whereas ``Oracle-prefix'' forces the corresponding ground-truth prefix before generating the remaining tokens. $\Delta$ indicates the absolute improvement over free decoding, with darker red signifying larger gains. Representative results are shown in \cref{tab:oracle_prefix_combined}, and full results across all datasets are provided in \cref{tab:oracle_prefix_combined_full}.

    \item \textbf{Distributional alignment.} The alignment between TIGER prefix generation and SASRec bucket-aggregated scoring on Beauty is shown in \cref{fig:beauty_js_pearson_combined}, with full numerical results provided in \cref{tab:bucket_alignment}.

    \item \textbf{TIGER-based diagnostic variants.} 
    We evaluate the seen and unseen recommendation performance of three TIGER-based variants under temporal cold-start settings. Representative results are shown in \cref{tab:variant_seen_unseen}, while full results are provided in \cref{tab:variant_seen_unseen_full}.

    \item \textbf{Variant token-level coldness taxonomy.} 
    This analysis presents the token-level coldness taxonomy for unseen target items across each variant. Representative results are shown in \cref{tab:variant_token_support}, with full results available in \cref{tab:variant_token_support_full}.
\end{itemize}

\begin{table*}[t]
\centering
\caption{Detailed @5 performance of TIGER under temporal cold-start evaluation.}
\label{tab:tiger_seen_unseen_at5}
\resizebox{0.6\textwidth}{!}{%
\begin{tabular}{l c ccc ccc}
\toprule
\multirow{2}{*}{\textbf{Dataset}}
& \multirow{2}{*}{\textbf{Cold ratio}}
& \multicolumn{3}{c}{\textbf{Seen}}
& \multicolumn{3}{c}{\textbf{Unseen}} \\
\cmidrule(lr){3-5}\cmidrule(lr){6-8}
&
& H@5 & R@5 & N@5
& H@5 & R@5 & N@5 \\
\midrule

\multicolumn{8}{l}{\textbf{Test Window $W_3$}} \\
\midrule
Beauty      & 0.984& \z & \z & \z & \z & \z & \z \\
Sports      & 1.000& -- & -- & -- & \z & \z & \z \\
Toys        & 0.903& \z & \z & \z & \z & \z & \z \\
\midrule
Sephora     & 0.634& 0.16915 & 0.19306 & 0.13842 & \z & \z & \z \\
WikiLife    & 0.650& 0.38298 & 0.35000 & 0.20709 & \z & \z & \z \\
WeiboDaily  & 0.285& 0.59873 & 0.22879 & 0.17338 & \z & \z & \z \\
WeiboTech   & 0.469& 0.23171 & 0.00243 & 0.00097 & \z & \z & \z \\
\midrule

\multicolumn{8}{l}{\textbf{Test Window $W_4$}} \\
\midrule
Beauty      & 0.957& 0.02439 & \z & \z & \z & \z & \z \\
Sports      & 0.960& \z & \z & \z & \z & \z & \z \\
Toys        & 0.901& 0.02703 & 0.00557 & 0.00240 & \z & \z & \z \\
\midrule
Sephora     & 0.823& 0.09255 & 0.14625 & 0.11730 & \z & \z & \z \\
WikiLife    & 0.593& 0.28244 & 0.18613 & 0.13446 & \z & \z & \z \\
WeiboDaily  & 0.247& 0.70188 & 0.28315 & 0.21476 & \z & \z & \z \\
WeiboTech   & 0.409& 0.27933 & 0.02312 & 0.01362 & \z & \z & \z \\
\midrule

\multicolumn{8}{l}{\textbf{Test Window $W_5$}} \\
\midrule
Beauty      & 0.904& 0.00504 & 0.00172 & 0.00117 & \z & \z & \z \\
Sports      & 0.883& 0.02145 & 0.00651 & 0.00424 & 0.00060 & 0.00003 & 0.00003 \\
Toys        & 0.925& 0.01038 & 0.00385 & 0.00276 & \z & \z & \z \\
\midrule
Sephora     & 0.863& 0.09608 & 0.19850 & 0.16848 & \z & \z & \z \\
WikiLife    & 0.463& 0.26678 & 0.20218 & 0.14697 & \z & \z & \z \\
WeiboDaily  & 0.228& 0.71147 & 0.29337 & 0.22213 & 0.00159 & 0.00018 & 0.00008 \\
WeiboTech   & 0.277& 0.41813 & 0.05636 & 0.03755 & \z & \z & \z \\
\midrule

\multicolumn{8}{l}{\textbf{Average}} \\
\midrule
Beauty      & 0.948& 0.00981 & 0.00057 & 0.00039 & \z & \z & \z \\
Sports      & 0.947& 0.01072 & 0.00326 & 0.00212 & 0.00020 & 0.00001 & 0.00001 \\
Toys        & 0.910& 0.01247 & 0.00314 & 0.00172 & \z & \z & \z \\
\midrule
Sephora     & 0.773& 0.11926 & 0.17927 & 0.14140 & \z & \z & \z \\
WikiLife    & 0.569& 0.31073 & 0.24610 & 0.16284 & \z & \z & \z \\
WeiboDaily  & 0.253& 0.67069 & 0.26844 & 0.20342 & 0.00053 & 0.00006 & 0.00003 \\
WeiboTech   & 0.385 & 0.30972 & 0.02730 & 0.01738 & \z & \z & \z \\

\bottomrule
\end{tabular}%
}
\end{table*}

\begin{table*}[t]
\centering
\caption{Detailed @5 performance of SASRec under temporal cold-start evaluation.}
\label{tab:sasrec_seen_unseen_at5}
\resizebox{0.6\textwidth}{!}{%
\begin{tabular}{l c ccc ccc}
\toprule
\multirow{2}{*}{\textbf{Dataset}}
& \multirow{2}{*}{\textbf{Cold ratio}}
& \multicolumn{3}{c}{\textbf{Seen}}
& \multicolumn{3}{c}{\textbf{Unseen}} \\
\cmidrule(lr){3-5}\cmidrule(lr){6-8}
&
& H@5 & R@5 & N@5
& H@5 & R@5 & N@5 \\
\midrule

\multicolumn{8}{l}{\textbf{Test Window $W_3$}} \\
\midrule
Beauty      & 0.984 & \z & \z & \z & \z & \z & \z \\
Sports      & 1.000 & -- & -- & -- & \z & \z & \z \\
Toys        & 0.903 & 0.01493 & 0.04598 & 0.03188 & \z & \z & \z \\
\midrule
Sephora     & 0.634 & 0.04736 & 0.03374 & 0.01363 & \z & \z & \z \\
WikiLife    & 0.650 & \z & 0.11000 & 0.08042 & \z & \z & \z \\
WeiboDaily  & 0.285 & 0.20255 & 0.36725 & 0.30346 & \z & \z & \z \\
WeiboTech   & 0.469 & 0.15854 & 0.04761 & 0.03131 & \z & \z & \z \\
\midrule

\multicolumn{8}{l}{\textbf{Test Window $W_4$}} \\
\midrule
Beauty      & 0.957 & \z & 0.00813 & 0.00513 & \z & \z & \z \\
Sports      & 0.960 & 0.02410 & 0.01042 & 0.01042 & 0.00157 & 0.00044 & 0.00019 \\
Toys        & 0.901 & 0.02252 & 0.00836 & 0.00403 & \z & \z & \z \\
\midrule
Sephora     & 0.823 & 0.04808 & 0.03557 & 0.01594 & 0.00126 & 0.00051 & 0.00026 \\
WikiLife    & 0.593 & 0.01145 & 0.00730 & 0.00571 & \z & \z & \z \\
WeiboDaily  & 0.247 & 0.23993 & 0.30098 & 0.24461 & \z & \z & \z \\
WeiboTech   & 0.409 & 0.15642 & 0.03122 & 0.01987 & \z & \z & \z \\
\midrule

\multicolumn{8}{l}{\textbf{Test Window $W_5$}} \\
\midrule
Beauty      & 0.904 & \z & 0.00086 & 0.00041 & 0.00046 & 0.00006 & 0.00003 \\
Sports      & 0.883 & 0.01125 & 0.00546 & 0.00338 & 0.00239 & 0.00006 & 0.00003 \\
Toys        & 0.925 & 0.00445 & 0.00342 & 0.00211 & 0.00232 & 0.00003 & 0.00002 \\
\midrule
Sephora     & 0.863 & 0.01079 & 0.02880 & 0.02331 & 0.02556 & 0.00616 & 0.00385 \\
WikiLife    & 0.463 & 0.00452 & 0.00627 & 0.00443 & \z & \z & \z \\
WeiboDaily  & 0.228 & 0.24229 & 0.33069 & 0.27326 & \z & \z & \z \\
WeiboTech   & 0.277 & 0.14912 & 0.03530 & 0.02260 & \z & \z & \z \\
\midrule

\multicolumn{8}{l}{\textbf{Average}} \\
\midrule
Beauty      & 0.948 & \z & 0.00300 & 0.00185 & 0.00015 & 0.00002 & 0.00001 \\
Sports      & 0.947 & 0.01767 & 0.00794 & 0.00690 & 0.00132 & 0.00016 & 0.00007 \\
Toys        & 0.910 & 0.01397 & 0.01925 & 0.01268 & 0.00077 & 0.00001 & 0.00001 \\
\midrule
Sephora     & 0.773 & 0.03541 & 0.03270 & 0.01763 & 0.00894 & 0.00222 & 0.00137 \\
WikiLife    & 0.569 & 0.00532 & 0.04119 & 0.03019 & \z & \z & \z \\
WeiboDaily  & 0.253 & 0.22826 & 0.33298 & 0.27378 & \z & \z & \z \\
WeiboTech   & 0.385 & 0.15469 & 0.03804 & 0.02459 & \z & \z & \z \\

\bottomrule
\end{tabular}%
}
\end{table*}

\begin{table*}[t]
\centering
\caption{Detailed @10 performance of TIGER under temporal cold-start evaluation.}
\label{tab:tiger_seen_unseen_at10}
\resizebox{0.6\textwidth}{!}{%
\begin{tabular}{l c ccc ccc}
\toprule
\multirow{2}{*}{\textbf{Dataset}}
& \multirow{2}{*}{\textbf{Cold ratio}}
& \multicolumn{3}{c}{\textbf{Seen}}
& \multicolumn{3}{c}{\textbf{Unseen}} \\
\cmidrule(lr){3-5}\cmidrule(lr){6-8}
&
& H@10 & R@10 & N@10
& H@10 & R@10 & N@10 \\
\midrule

\multicolumn{8}{l}{\textbf{Test Window $W_3$}} \\
\midrule
Beauty      & 0.984 & \z & \z & \z & \z & \z & \z \\
Sports      & 1.000 & -- & -- & -- & \z & \z & \z \\
Toys        & 0.903 & \z & \z & \z & \z & \z & \z \\
\midrule
Sephora     & 0.634 & 0.28281 & 0.27648 & 0.16501 & \z & \z & \z \\
WikiLife    & 0.650 & 0.38298 & 0.41000 & 0.22805 & \z & \z & \z \\
WeiboDaily  & 0.285 & 0.65860 & 0.28977 & 0.19304 & \z & \z & \z \\
WeiboTech   & 0.469 & 0.25610 & 0.00932 & 0.00315 & \z & \z & \z \\

\midrule
\multicolumn{8}{l}{\textbf{Test Window $W_4$}} \\
\midrule
Beauty      & 0.957 & 0.02439 & \z & \z & \z & \z & \z \\
Sports      & 0.960 & \z & \z & \z & \z & \z & \z \\
Toys        & 0.901 & 0.03153 & 0.01393 & 0.00531 & \z & \z & \z \\
\midrule
Sephora     & 0.823 & 0.18029 & 0.21621 & 0.13985 & \z & \z & \z \\
WikiLife    & 0.593 & 0.35878 & 0.21715 & 0.14444 & \z & \z & \z \\
WeiboDaily  & 0.247 & 0.74306 & 0.35124 & 0.23682 & \z & 0.00041 & 0.00013 \\
WeiboTech   & 0.409 & 0.36313 & 0.03776 & 0.01831 & \z & \z & \z \\

\midrule
\multicolumn{8}{l}{\textbf{Test Window $W_5$}} \\
\midrule
Beauty      & 0.904 & 0.01008 & 0.00459 & 0.00209 & 0.00023 & \z & \z \\
Sports      & 0.883 & 0.02743 & 0.01155 & 0.00580 & 0.00060 & 0.00006 & 0.00004 \\
Toys        & 0.925 & 0.01631 & 0.00685 & 0.00373 & \z & \z & \z \\
\midrule
Sephora     & 0.863 & 0.17361 & 0.24636 & 0.18382 & \z & \z & \z \\
WikiLife    & 0.463 & 0.34122 & 0.26803 & 0.16834 & \z & \z & \z \\
WeiboDaily  & 0.228 & 0.75832 & 0.36982 & 0.24692 & 0.00319 & 0.00044 & 0.00017 \\
WeiboTech   & 0.277 & 0.50585 & 0.08658 & 0.04726 & \z & \z & \z \\

\midrule
\multicolumn{8}{l}{\textbf{Average}} \\
\midrule
Beauty      & 0.948 & 0.01149 & 0.00153 & 0.00070 & 0.00008 & \z & \z \\
Sports      & 0.947 & 0.01371 & 0.00578 & 0.00290 & 0.00020 & 0.00002 & 0.00001 \\
Toys        & 0.910 & 0.01595 & 0.00692 & 0.00301 & \z & \z & \z \\
\midrule
Sephora     & 0.773 & 0.21224 & 0.24635 & 0.16289 & \z & \z & \z \\
WikiLife    & 0.569 & 0.36099 & 0.29840 & 0.18028 & \z & \z & \z \\
WeiboDaily  & 0.253 & 0.71999 & 0.33694 & 0.22559 & 0.00106 & 0.00028 & 0.00010 \\
WeiboTech   & 0.385 & 0.37502 & 0.04455 & 0.02291 & \z & \z & \z \\

\bottomrule
\end{tabular}%
}
\end{table*}

\begin{table*}[t]
\centering
\caption{Detailed @10 performance of SASRec under temporal cold-start evaluation.}
\label{tab:sasrec_seen_unseen_at10}
\resizebox{0.6\textwidth}{!}{%
\begin{tabular}{l c ccc ccc}
\toprule
\multirow{2}{*}{\textbf{Dataset}}
& \multirow{2}{*}{\textbf{Cold ratio}}
& \multicolumn{3}{c}{\textbf{Seen}}
& \multicolumn{3}{c}{\textbf{Unseen}} \\
\cmidrule(lr){3-5}\cmidrule(lr){6-8}
&
& H@10 & R@10 & N@10
& H@10 & R@10 & N@10 \\
\midrule

\multicolumn{8}{l}{\textbf{Test Window $W_3$}} \\
\midrule
Beauty      & 0.984 & \z & \z & \z & \z & \z & \z \\
Sports      & 1.000 & -- & -- & -- & \z & \z & \z \\
Toys        & 0.903 & 0.05970 & 0.06897 & 0.03904 & \z & \z & \z \\
\midrule
Sephora     & 0.634 & 0.04736 & 0.06654 & 0.02531 & 0.06304 & 0.03568 & 0.01144 \\
WikiLife    & 0.650 & 0.04255 & 0.14000 & 0.09110 & \z & \z & \z \\
WeiboDaily  & 0.285 & 0.22166 & 0.42399 & 0.32175 & \z & \z & \z \\
WeiboTech   & 0.469 & 0.28049 & 0.07224 & 0.03926 & \z & \z & \z \\

\midrule
\multicolumn{8}{l}{\textbf{Test Window $W_4$}} \\
\midrule
Beauty      & 0.957 & \z & 0.02439 & 0.01014 & \z & \z & \z \\
Sports      & 0.960 & 0.02410 & 0.07292 & 0.03017 & 0.00315 & 0.00044 & 0.00019 \\
Toys        & 0.901 & 0.04505 & 0.03343 & 0.01213 & \z & \z & \z \\
\midrule
Sephora     & 0.823 & 0.06971 & 0.08182 & 0.03205 & 0.01912 & 0.00800 & 0.00259 \\
WikiLife    & 0.593 & 0.01527 & 0.00912 & 0.00636 & \z & \z & \z \\
WeiboDaily  & 0.247 & 0.25067 & 0.36107 & 0.26405 & 0.00207 & \z & \z \\
WeiboTech   & 0.409 & 0.23464 & 0.05053 & 0.02614 & \z & \z & \z \\

\midrule
\multicolumn{8}{l}{\textbf{Test Window $W_5$}} \\
\midrule
Beauty      & 0.904 & 0.00050 & 0.00201 & 0.00077 & 0.00138 & 0.00018 & 0.00007 \\
Sports      & 0.883 & 0.02286 & 0.00987 & 0.00483 & 0.00582 & 0.00011 & 0.00004 \\
Toys        & 0.925 & 0.00741 & 0.00685 & 0.00323 & 0.00336 & 0.00010 & 0.00005 \\
\midrule
Sephora     & 0.863 & 0.07403 & 0.07217 & 0.03721 & 0.04006 & 0.01476 & 0.00638 \\
WikiLife    & 0.463 & 0.00557 & 0.00996 & 0.00563 & 0.00039 & 0.00043 & 0.00014 \\
WeiboDaily  & 0.228 & 0.26572 & 0.38945 & 0.29224 & 0.00159 & 0.00009 & 0.00003 \\
WeiboTech   & 0.277 & 0.22515 & 0.05845 & 0.03002 & \z & \z & \z \\

\midrule
\multicolumn{8}{l}{\textbf{Average}} \\
\midrule
Beauty      & 0.948 & 0.00017 & 0.00880 & 0.00364 & 0.00046 & 0.00006 & 0.00002 \\
Sports      & 0.947 & 0.02348 & 0.04140 & 0.01750 & 0.00299 & 0.00018 & 0.00008 \\
Toys        & 0.910 & 0.03739 & 0.03641 & 0.01813 & 0.00112 & 0.00003 & 0.00002 \\
\midrule
Sephora     & 0.773 & 0.06370 & 0.07351 & 0.03152 & 0.04074 & 0.01948 & 0.00681 \\
WikiLife    & 0.569 & 0.02113 & 0.05303 & 0.03437 & 0.00013 & 0.00014 & 0.00005 \\
WeiboDaily  & 0.253 & 0.24602 & 0.39150 & 0.29268 & 0.00122 & 0.00003 & 0.00001 \\
WeiboTech   & 0.385 & 0.24676 & 0.06041 & 0.03181 & \z & \z & \z \\

\bottomrule
\end{tabular}%
}
\end{table*}

\begin{table*}[t]
\centering
\caption{Detailed @20 performance of TIGER under temporal cold-start evaluation.}
\label{tab:tiger_seen_unseen_at20}
\resizebox{0.6\textwidth}{!}{%
\begin{tabular}{l c ccc ccc}
\toprule
\multirow{2}{*}{\textbf{Dataset}}
& \multirow{2}{*}{\textbf{Cold ratio}}
& \multicolumn{3}{c}{\textbf{Seen}}
& \multicolumn{3}{c}{\textbf{Unseen}} \\
\cmidrule(lr){3-5}\cmidrule(lr){6-8}
&
& H@20 & R@20 & N@20
& H@20 & R@20 & N@20 \\
\midrule

\multicolumn{8}{l}{\textbf{Test Window $W_3$}} \\
\midrule
Beauty      & 0.984 & \z & \z & \z & \z & \z & \z \\
Sports      & 1.000 & -- & -- & -- & \z & \z & \z \\
Toys        & 0.903 & \z & \z & \z & \z & \z & \z \\
\midrule
Sephora     & 0.634 & 0.43031 & 0.37676 & 0.19033 & \z & \z & \z \\
WikiLife    & 0.650 & 0.38298 & 0.41000 & 0.22805 & \z & \z & \z \\
WeiboDaily  & 0.285 & 0.70318 & 0.35125 & 0.20866 & 0.00253 & \z & \z \\
WeiboTech   & 0.469 & 0.25610 & 0.01378 & 0.00428 & \z & \z & \z \\

\midrule
\multicolumn{8}{l}{\textbf{Test Window $W_4$}} \\
\midrule
Beauty      & 0.957 & 0.02439 & \z & \z & \z & \z & \z \\
Sports      & 0.960 & \z & \z & \z & \z & \z & \z \\
Toys        & 0.901 & 0.04054 & 0.02786 & 0.00884 & \z & \z & \z \\
\midrule
Sephora     & 0.823 & 0.30288 & 0.31581 & 0.16471 & \z & \z & \z \\
WikiLife    & 0.593 & 0.38168 & 0.28285 & 0.16094 & 0.00310 & 0.00125 & 0.00028 \\
WeiboDaily  & 0.247 & 0.77708 & 0.41788 & 0.25367 & 0.00207 & 0.00101 & 0.00028 \\
WeiboTech   & 0.409 & 0.40782 & 0.05434 & 0.02248 & \z & \z & \z \\

\midrule
\multicolumn{8}{l}{\textbf{Test Window $W_5$}} \\
\midrule
Beauty      & 0.904 & 0.01562 & 0.00775 & 0.00288 & 0.00069 & \z & \z \\
Sports      & 0.883 & 0.02813 & 0.01702 & 0.00720 & 0.00060 & 0.00011 & 0.00005 \\
Toys        & 0.925 & 0.02965 & 0.01326 & 0.00534 & 0.00026 & \z & \z \\
\midrule
Sephora     & 0.863 & 0.27835 & 0.31991 & 0.20237 & \z & \z & \z \\
WikiLife    & 0.463 & 0.40870 & 0.32319 & 0.18234 & \z & \z & \z \\
WeiboDaily  & 0.228 & 0.79655 & 0.43705 & 0.26397 & 0.00478 & 0.00176 & 0.00051 \\
WeiboTech   & 0.277 & 0.56725 & 0.13624 & 0.05972 & \z & \z & \z \\

\midrule
\multicolumn{8}{l}{\textbf{Average}} \\
\midrule
Beauty      & 0.948 & 0.01334 & 0.00258 & 0.00096 & 0.00023 & \z & \z \\
Sports      & 0.947 & 0.01406 & 0.00851 & 0.00360 & 0.00020 & 0.00004 & 0.00002 \\
Toys        & 0.910 & 0.02340 & 0.01371 & 0.00473 & 0.00009 & \z & \z \\
\midrule
Sephora     & 0.773 & 0.33718 & 0.33749 & 0.18581 & \z & \z & \z \\
WikiLife    & 0.569 & 0.39112 & 0.33868 & 0.19044 & 0.00103 & 0.00042 & 0.00009 \\
WeiboDaily  & 0.253 & 0.75894 & 0.40206 & 0.24210 & 0.00313 & 0.00092 & 0.00026 \\
WeiboTech   & 0.385 & 0.41039 & 0.06812 & 0.02883 & \z & \z & \z \\

\bottomrule
\end{tabular}%
}
\end{table*}

\begin{table*}[t]
\centering
\caption{Detailed @20 performance of SASRec under temporal cold-start evaluation.}
\label{tab:sasrec_seen_unseen_at20}
\resizebox{0.6\textwidth}{!}{%
\begin{tabular}{l c ccc ccc}
\toprule
\multirow{2}{*}{\textbf{Dataset}}
& \multirow{2}{*}{\textbf{Cold ratio}}
& \multicolumn{3}{c}{\textbf{Seen}}
& \multicolumn{3}{c}{\textbf{Unseen}} \\
\cmidrule(lr){3-5}\cmidrule(lr){6-8}
&
& H@20 & R@20 & N@20
& H@20 & R@20 & N@20 \\
\midrule

\multicolumn{8}{l}{\textbf{Test Window $W_3$}} \\
\midrule
Beauty      & 0.984 & 0.25000 & \z & \z & \z & \z & \z \\
Sports      & 1.000 & -- & -- & -- & \z & \z & \z \\
Toys        & 0.903 & 0.08955 & 0.10345 & 0.04738 & \z & \z & \z \\
\midrule
Sephora     & 0.634 & 0.06089 & 0.08529 & 0.02965 & 0.06304 & 0.03568 & 0.01144 \\
WikiLife    & 0.650 & 0.06383 & 0.15000 & 0.09355 & 0.03448 & 0.03226 & 0.00759 \\
WeiboDaily  & 0.285 & 0.25096 & 0.48114 & 0.33620 & \z & \z & \z \\
WeiboTech   & 0.469 & 0.43902 & 0.10504 & 0.04755 & \z & \z & \z \\

\midrule
\multicolumn{8}{l}{\textbf{Test Window $W_4$}} \\
\midrule
Beauty      & 0.957 & \z & 0.03252 & 0.01206 & 0.00253 & 0.00074 & 0.00018 \\
Sports      & 0.960 & 0.07229 & 0.12500 & 0.04304 & 0.00472 & 0.00044 & 0.00019 \\
Toys        & 0.901 & 0.05405 & 0.05014 & 0.01638 & \z & \z & \z \\
\midrule
Sephora     & 0.823 & 0.08053 & 0.09960 & 0.03622 & 0.03571 & 0.01448 & 0.00423 \\
WikiLife    & 0.593 & 0.04580 & 0.03467 & 0.01253 & \z & \z & \z \\
WeiboDaily  & 0.247 & 0.26321 & 0.42386 & 0.27989 & 0.00414 & \z & \z \\
WeiboTech   & 0.409 & 0.28492 & 0.07552 & 0.03244 & \z & \z & \z \\

\midrule
\multicolumn{8}{l}{\textbf{Test Window $W_5$}} \\
\midrule
Beauty      & 0.904 & 0.00101 & 0.00258 & 0.00092 & 0.00988 & 0.00129 & 0.00035 \\
Sports      & 0.883 & 0.02707 & 0.01387 & 0.00584 & 0.00776 & 0.00022 & 0.00007 \\
Toys        & 0.925 & 0.01334 & 0.01412 & 0.00503 & 0.00439 & 0.00010 & 0.00005 \\
\midrule
Sephora     & 0.863 & 0.07875 & 0.07743 & 0.03846 & 0.07120 & 0.02582 & 0.00917 \\
WikiLife    & 0.463 & 0.01148 & 0.01568 & 0.00704 & 0.00039 & 0.00043 & 0.00014 \\
WeiboDaily  & 0.228 & 0.30086 & 0.45330 & 0.30837 & 0.00159 & 0.00009 & 0.00003 \\
WeiboTech   & 0.277 & 0.27778 & 0.09432 & 0.03901 & \z & \z & \z \\

\midrule
\multicolumn{8}{l}{\textbf{Average}} \\
\midrule
Beauty      & 0.948 & 0.08367 & 0.01170 & 0.00433 & 0.00414 & 0.00068 & 0.00018 \\
Sports      & 0.947 & 0.04968 & 0.06943 & 0.02444 & 0.00416 & 0.00022 & 0.00009 \\
Toys        & 0.910 & 0.05232 & 0.05590 & 0.02293 & 0.00146 & 0.00003 & 0.00002 \\
\midrule
Sephora     & 0.773 & 0.07339 & 0.08744 & 0.03478 & 0.05665 & 0.02532 & 0.00828 \\
WikiLife    & 0.569 & 0.04037 & 0.06678 & 0.03771 & 0.01163 & 0.01090 & 0.00258 \\
WeiboDaily  & 0.253 & 0.27167 & 0.45276 & 0.30815 & 0.00191 & 0.00003 & 0.00001 \\
WeiboTech   & 0.385 & 0.33391 & 0.09163 & 0.03967 & \z & \z & \z \\

\bottomrule
\end{tabular}%
}
\end{table*}

\begin{table*}[t]
\centering
\caption{Token-level coldness taxonomy for unseen target items. The table separates cold items according to token support and prefix support.}
\label{tab:coldness_breakdown_full}
\resizebox{0.85\textwidth}{!}{
\begin{tabular}{l l c ccc ccc}
\toprule
\multirow{2}{*}{\textbf{Dataset}}
& \multirow{2}{*}{\textbf{Coldness category}}
& \multirow{2}{*}{\textbf{Ratio}}
& \multicolumn{6}{c}{\textbf{TIGER}} \\
\cmidrule(lr){4-9}
&
&
& Recall@5 & Recall@10 & Recall@20
& NDCG@5 & NDCG@10 & NDCG@20 \\
\midrule

\multirow{5}{*}{Beauty}
& all-token-seen               & 0.521 & \z & \z & \z & \z & \z & \z \\
& any-token-unseen             & 0.479 & \z & \z & \z & \z & \z & \z \\
& prefix-seen-$c_1$            & 0.954 & \z & \z & \z & \z & \z & \z \\
& prefix-seen-$c_1,c_2$        & 0.173 & \z & \z & \z & \z & \z & \z \\
& prefix-seen-$c_1,c_2,c_3$    & 0.001 & \z & \z & \z & \z & \z & \z \\
\midrule

\multirow{5}{*}{Sports}
& all-token-seen               & 0.405 & 0.00001 & 0.00003 & 0.00006 & 0.00001 & 0.00002 & 0.00003 \\
& any-token-unseen             & 0.595 & \z & \z & \z & \z & \z & \z \\
& prefix-seen-$c_1$            & 0.723 & 0.00001 & 0.00002 & 0.00004 & 0.00001 & 0.00001 & 0.00002 \\
& prefix-seen-$c_1,c_2$        & 0.208 & 0.00002 & 0.00004 & 0.00008 & 0.00002 & 0.00003 & 0.00004 \\
& prefix-seen-$c_1,c_2,c_3$    & 0.002 & 0.00336 & 0.00671 & 0.01007 & 0.00336 & 0.00437 & 0.00520 \\
\midrule

\multirow{5}{*}{Toys}
& all-token-seen               & 0.713 & \z & \z & \z & \z & \z & \z \\
& any-token-unseen             & 0.287 & \z & \z & \z & \z & \z & \z \\
& prefix-seen-$c_1$            & 1.000 & \z & \z & \z & \z & \z & \z \\
& prefix-seen-$c_1,c_2$        & 0.452 & \z & \z & \z & \z & \z & \z \\
& prefix-seen-$c_1,c_2,c_3$    & 0.004 & \z & \z & \z & \z & \z & \z \\
\midrule

\multirow{5}{*}{Sephora}
& all-token-seen               & 0.475 & \z & \z & \z & \z & \z & \z \\
& any-token-unseen             & 0.525 & \z & \z & \z & \z & \z & \z \\
& prefix-seen-$c_1$            & 1.000 & \z & \z & \z & \z & \z & \z \\
& prefix-seen-$c_1,c_2$        & 0.650 & \z & \z & \z & \z & \z & \z \\
& prefix-seen-$c_1,c_2,c_3$    & 0.002 & \z & \z & \z & \z & \z & \z \\
\midrule

\multirow{5}{*}{WikiLife}
& all-token-seen               & 0.824 & \z & \z & 0.00045 & \z & \z & 0.00010 \\
& any-token-unseen             & 0.176 & \z & \z & \z & \z & \z & \z \\
& prefix-seen-$c_1$            & 0.977 & \z & \z & 0.00042 & \z & \z & 0.00010 \\
& prefix-seen-$c_1,c_2$        & 0.304 & \z & \z & 0.00175 & \z & \z & 0.00040 \\
& prefix-seen-$c_1,c_2,c_3$    & 0.001 & \z & \z & \z & \z & \z & \z \\
\midrule

\multirow{5}{*}{WeiboDaily}
& all-token-seen               & 1.000 & 0.00006 & 0.00028 & 0.00092 & 0.00003 & 0.00010 & 0.00026 \\
& any-token-unseen             & 0.000 & -- & -- & -- & -- & -- & -- \\
& prefix-seen-$c_1$            & 1.000 & 0.00006 & 0.00028 & 0.00092 & 0.00003 & 0.00010 & 0.00026 \\
& prefix-seen-$c_1,c_2$        & 0.516 & 0.00009 & 0.00038 & 0.00138 & 0.00004 & 0.00014 & 0.00039 \\
& prefix-seen-$c_1,c_2,c_3$    & 0.010 & 0.00080 & 0.00320 & 0.01386 & 0.00034 & 0.00114 & 0.00379 \\
\midrule

\multirow{5}{*}{WeiboTech}
& all-token-seen               & 0.999 & \z & \z & \z & \z & \z & \z \\
& any-token-unseen             & 0.001 & \z & \z & \z & \z & \z & \z \\
& prefix-seen-$c_1$            & 0.999 & \z & \z & \z & \z & \z & \z \\
& prefix-seen-$c_1,c_2$        & 0.591 & \z & \z & \z & \z & \z & \z \\
& prefix-seen-$c_1,c_2,c_3$    & 0.008 & \z & \z & \z & \z & \z & \z \\

\bottomrule
\end{tabular}}
\end{table*}

\begin{table*}[t]
\centering
\caption{Oracle-prefix probing for semantic ID generation.
We report both next-item semantic ID generation accuracy and next-token semantic ID generation accuracy.
``Free'' denotes standard autoregressive decoding, while ``Oracle-prefix'' decoding forces the corresponding ground-truth prefix before generating the remaining tokens.
$\Delta$ denotes the absolute improvement over free decoding. Darker red indicates larger improvement.}
\label{tab:oracle_prefix_combined_full}
\resizebox{\textwidth}{!}{
\begin{tabular}{l l c c cc cc c ccc ccc}
\toprule
\multirow{3}{*}{\textbf{Dataset}}
& \multirow{3}{*}{\textbf{Group}}
& \multirow{3}{*}{\textbf{Ratio}}
& \multicolumn{5}{c}{\textbf{Next-item SID generation}}
& \multicolumn{7}{c}{\textbf{Next-token SID generation}} \\
\cmidrule(lr){4-8}\cmidrule(lr){9-15}
&
&
& \multirow{2}{*}{\textbf{Free}}
& \multicolumn{2}{c}{\textbf{Oracle $c_1$}}
& \multicolumn{2}{c}{\textbf{Oracle $c_1,c_2$}}
& \multicolumn{1}{c}{\textbf{$c_1$ Acc.}}
& \multicolumn{3}{c}{\textbf{$c_2$ Acc.}}
& \multicolumn{3}{c}{\textbf{$c_3$ Acc.}} \\
\cmidrule(lr){5-6}\cmidrule(lr){7-8}
\cmidrule(lr){9-9}\cmidrule(lr){10-12}\cmidrule(lr){13-15}
&
&
&
& Acc. & $\Delta$
& Acc. & $\Delta$
& Free
& Free & Oracle $c_1$ & $\Delta$
& Free & Oracle $c_1,c_2$ & $\Delta$ \\
\midrule

\multirow{2}{*}{Beauty}
& Seen   & 0.052 & 0.00010 & 0.00823 & \dlight{0.00813} & 0.02383 & \dmid{0.02373} & 0.07487 & 0.02465 & 0.03639 & \dmid{0.01174} & 0.00163 & 0.02421 & \dmid{0.02259} \\
& Unseen & 0.948 & \z & \z & \dlight{\z} & 0.00134 & \dlight{0.00134} & 0.02181 & 0.00778 & 0.00870 & \dlight{0.00092} & 0.00126 & 0.00140 & \dlight{0.00014} \\
\midrule

\multirow{2}{*}{Sports}
& Seen   & 0.053 & 0.00063 & 0.00630 & \dlight{0.00567} & 0.02815 & \dmid{0.02752} & 0.18009 & 0.04449 & 0.07936 & \dmid{0.03487} & 0.00284 & 0.02910 & \dmid{0.02626} \\
& Unseen & 0.947 & 0.00001 & 0.00006 & \dlight{0.00005} & 0.00448 & \dlight{0.00447} & 0.11983 & 0.00494 & 0.01091 & \dlight{0.00597} & 0.00138 & 0.00530 & \dlight{0.00392} \\
\midrule

\multirow{2}{*}{Toys}
& Seen   & 0.090 & 0.00057 & 0.00564 & \dlight{0.00506} & 0.09664 & \dmid{0.09607} & 0.25815 & 0.01563 & 0.04140 & \dmid{0.02577} & 0.00514 & 0.09814 & \dmid{0.09300} \\
& Unseen & 0.910 & \z & \z & \dlight{\z} & 0.00272 & \dlight{0.00272} & 0.19895 & 0.00587 & 0.01855 & \dmid{0.01268} & 0.00391 & 0.00334 & -0.00057 \\
\midrule

\multirow{2}{*}{Sephora}
& Seen   & 0.227 & 0.10155 & 0.15851 & \dmid{0.05696} & 0.69089 & \ddeep{0.58933} & 0.38365 & 0.12973 & 0.21057 & \dmid{0.08084} & 0.11470 & 0.69117 & \ddeep{0.57648} \\
& Unseen & 0.773 & \z & \z & \dlight{\z} & 0.00063 & \dlight{0.00063} & 0.29701 & 0.01904 & 0.05075 & \dmid{0.03171} & 0.00679 & 0.00072 & -0.00606 \\
\midrule

\multirow{2}{*}{WikiLife}
& Seen   & 0.431 & 0.09446 & 0.16480 & \dmid{0.07034} & 0.50373 & \ddeep{0.40927} & 0.53002 & 0.12913 & 0.18835 & \dmid{0.05922} & 0.09624 & 0.50379 & \ddeep{0.40755} \\
& Unseen & 0.569 & \z & \z & \dlight{\z} & 0.00308 & \dlight{0.00308} & 0.27493 & 0.00845 & 0.01816 & \dlight{0.00971} & 0.00384 & 0.00322 & -0.00062 \\
\midrule

\multirow{2}{*}{WeiboDaily}
& Seen   & 0.747 & 0.12319 & 0.47262 & \ddeep{0.34943} & 0.75122 & \ddeep{0.62804} & 0.18646 & 0.13814 & 0.49332 & \ddeep{0.35518} & 0.12957 & 0.75163 & \ddeep{0.62206} \\
& Unseen & 0.253 & 0.00003 & 0.00049 & \dlight{0.00046} & 0.01493 & \dmid{0.01490} & 0.06264 & 0.01797 & 0.02774 & \dlight{0.00977} & 0.01480 & 0.01972 & \dlight{0.00492} \\
\midrule

\multirow{2}{*}{WeiboTech}
& Seen   & 0.615 & 0.00632 & 0.07472 & \dmid{0.06841} & 0.39882 & \ddeep{0.39251} & 0.08608 & 0.01498 & 0.10570 & \dmid{0.09071} & 0.01139 & 0.39933 & \ddeep{0.38794} \\
& Unseen & 0.385 & \z & 0.00004 & \dlight{0.00004} & 0.01173 & \dmid{0.01173} & 0.06236 & 0.00881 & 0.01749 & \dlight{0.00868} & 0.00647 & 0.01666 & \dmid{0.01020} \\

\bottomrule
\end{tabular}}
\end{table*}

\begin{table*}[t]
\centering
\caption{Distributional alignment between TIGER prefix generation and SASRec bucket-aggregated scoring.
Each row compares the TIGER prefix distribution $P_T(\cdot)$ with the SASRec bucket distribution $P_S(\cdot)$ at the corresponding semantic-ID prefix level.
Lower JS divergence and higher Pearson correlation indicate stronger alignment.}
\label{tab:bucket_alignment}
\resizebox{0.75\textwidth}{!}{
\begin{tabular}{l l c c c c c c c}
\toprule
\textbf{Metric}
& \textbf{SID prefix}
& \textbf{Beauty}
& \textbf{Sports}
& \textbf{Toys}
& \textbf{Sephora}
& \textbf{WikiLife}
& \textbf{WeiboDaily}
& \textbf{WeiboTech} \\
\midrule
\multirow{3}{*}{JS divergence $\downarrow$}
& $c_1$         & 0.12821 & 0.29669 & 0.25995 & 0.09539 & 0.19576 & 0.20870 & 0.24888 \\
& $c_1,c_2$     & 0.49832 & 0.53745 & 0.56818 & 0.33208 & 0.36511 & 0.39852 & 0.56114 \\
& $c_1,c_2,c_3$ & 0.64622 & 0.63961 & 0.66619 & 0.47234 & 0.40976 & 0.46563 & 0.60842 \\
\midrule
\multirow{3}{*}{Pearson $\uparrow$}
& $c_1$         & 0.58776 & 0.46533 & 0.44644 & 0.63721 & 0.58386 & 0.53238 & 0.36585 \\
& $c_1,c_2$     & 0.39636 & 0.29548 & 0.18975 & 0.42604 & 0.39078 & 0.36548 & 0.11598 \\
& $c_1,c_2,c_3$ & 0.19580 & 0.00009 & 0.00480 & 0.29056 & 0.28029 & 0.16506 & 0.10909 \\
\bottomrule
\end{tabular}}
\end{table*}

% \begin{table*}[t]
% \centering
% \caption{Seen and unseen recommendation performance of TIGER-based diagnostic variants under temporal cold-start evaluation.}
% \label{tab:variant_seen_unseen_full}
% \resizebox{\textwidth}{!}{
% \begin{tabular}{l c cc cc cc cc}
% \toprule
% \multirow{2}{*}{\textbf{Dataset}}
% & \multirow{2}{*}{\textbf{Cold ratio}}
% & \multicolumn{2}{c}{\textbf{TIGER}}
% & \multicolumn{2}{c}{\textbf{TIGER-SID}}
% & \multicolumn{2}{c}{\textbf{TIGER-Scorer}}
% & \multicolumn{2}{c}{\textbf{TIGER-Edge}} \\
% \cmidrule(lr){3-4}\cmidrule(lr){5-6}\cmidrule(lr){7-8}\cmidrule(lr){9-10}
% &
% & Seen H@20 & Unseen H@20
% & Seen H@20 & Unseen H@20
% & Seen H@20 & Unseen H@20
% & Seen H@20 & Unseen H@20 \\
% \midrule
% Beauty      & 0.948 & 0.01334 & 0.00023 & 0.00655 & 0.00054 & 0.02163 & 0.00831 & 0.01925 & 0.00042 \\
% Sports      & 0.947 & 0.01406 & 0.00020 & 0.01582 & 0.00109 & 0.04120 & 0.01392 & 0.03886 & 0.00075 \\
% Toys        & 0.910 & 0.02340 & 0.00009 & 0.02790 & \z & 0.00920 & 0.01187 & 0.08387 & \z \\
% \midrule
% Sephora     & 0.773 & 0.33718 & \z & 0.32466 & \z & 0.24545 & 0.00003 & 0.31388 & 0.00198 \\
% WikiLife    & 0.569 & 0.39112 & 0.00103 & 0.38787 & 0.00013 & 0.28630 & 0.00505 & 0.43409 & 0.00026 \\
% WeiboDaily  & 0.253 & 0.75894 & 0.00313 & 0.77940 & 0.00747 & 0.32633 & 0.00750 & 0.81308 & 0.01132 \\
% WeiboTech   & 0.385 & 0.41039 & \z & 0.49199 & 0.01232 & 0.16490 & 0.02929 & 0.47153 & 0.01683 \\
% \bottomrule
% \end{tabular}}
% \end{table*}

\begin{table*}[t]
\centering
\caption{Seen and unseen recommendation performance of TIGER-based diagnostic variants under temporal cold-start evaluation. Arrows indicate relative change compared to TIGER (\textcolor{red}{$\uparrow$},\textcolor{blue}{$\downarrow$}: >100\%, \textcolor{red!40}{$\uparrow$},\textcolor{blue!40}{$\downarrow$}: $\le$100\%).}
\label{tab:variant_seen_unseen_full}
\resizebox{\textwidth}{!}{
\begin{tabular}{l c cc cc cc cc}
\toprule
\multirow{2}{*}{\textbf{Dataset}}
& \multirow{2}{*}{\textbf{Cold ratio}}
& \multicolumn{2}{c}{\textbf{TIGER}}
& \multicolumn{2}{c}{\textbf{TIGER-SID}}
& \multicolumn{2}{c}{\textbf{TIGER-Scorer}}
& \multicolumn{2}{c}{\textbf{TIGER-Edge}} \\
\cmidrule(lr){3-4}\cmidrule(lr){5-6}\cmidrule(lr){7-8}\cmidrule(lr){9-10}
&
& Seen H@20 & Unseen H@20
& Seen H@20 & Unseen H@20
& Seen H@20 & Unseen H@20
& Seen H@20 & Unseen H@20 \\

\midrule
Beauty      & 0.948 & 0.01334 & 0.00023 
& 0.00655\,\textcolor{blue!40}{$\downarrow$} & 0.00054\,\textcolor{red!40}{$\uparrow$} 
& \textbf{0.02163}\,\textcolor{red!40}{$\uparrow$} & \textbf{0.00831}\,\textcolor{red}{$\uparrow$} 
& {0.01925}\,\textcolor{red!40}{$\uparrow$} & 0.00042\,\textcolor{red!40}{$\uparrow$} \\

Sports      & 0.947& 0.01406 & 0.00020 
& 0.01582\,\textcolor{red!40}{$\uparrow$} & 0.00109\,\textcolor{red}{$\uparrow$} 
& \textbf{0.04120}\,\textcolor{red}{$\uparrow$} & \textbf{0.01392}\,\textcolor{red}{$\uparrow$} 
& 0.03886\,\textcolor{red}{$\uparrow$} & 0.00075\,\textcolor{red}{$\uparrow$} \\

Toys        & 0.910 & 0.02340 & 0.00009 
& 0.02790\,\textcolor{red!40}{$\uparrow$} & \z\,\textcolor{blue!40}{$\downarrow$} 
& 0.00920\,\textcolor{blue!40}{$\downarrow$} & \textbf{0.01187}\,\textcolor{red}{$\uparrow$} 
& \textbf{0.08387}\,\textcolor{red}{$\uparrow$} & \z\,\textcolor{blue!40}{$\downarrow$} \\

\midrule
Sephora     & 0.773 & \textbf{0.33718} & \z 
& 0.32466\,\textcolor{blue!40}{$\downarrow$} & \z\,\textcolor{blue!40}{$\downarrow$} 
& 0.24545\,\textcolor{blue!40}{$\downarrow$} & 0.00003\,\textcolor{red}{$\uparrow$} 
& 0.31388\,\textcolor{blue!40}{$\downarrow$} & \textbf{0.00198}\,\textcolor{red}{$\uparrow$} \\

WikiLife    & 0.569 & 0.39112 & 0.00103 
& 0.38787\,\textcolor{blue!40}{$\downarrow$} & 0.00013\,\textcolor{blue!40}{$\downarrow$} 
& 0.28630\,\textcolor{blue!40}{$\downarrow$} & \textbf{0.00505}\,\textcolor{red}{$\uparrow$} 
& \textbf{0.43409}\,\textcolor{red!40}{$\uparrow$} & 0.00026\,\textcolor{blue!40}{$\downarrow$} \\

WeiboDaily  & 0.253& 0.75894 & 0.00313 
& 0.77940\,\textcolor{red!40}{$\uparrow$} & 0.00747\,\textcolor{red}{$\uparrow$} 
& 0.32633\,\textcolor{blue!40}{$\downarrow$} & 0.00750\,\textcolor{red}{$\uparrow$} 
& \textbf{0.81308}\,\textcolor{red!40}{$\uparrow$} & \textbf{0.01132}\,\textcolor{red}{$\uparrow$} \\

WeiboTech   & 0.385& 0.41039 & \z 
& \textbf{0.49199}\,\textcolor{red!40}{$\uparrow$} & 0.01232\,\textcolor{red}{$\uparrow$} 
& 0.16490\,\textcolor{blue!40}{$\downarrow$} & \textbf{0.02929}\,\textcolor{red}{$\uparrow$} 
& {0.47153}\,\textcolor{red!40}{$\uparrow$} & 0.01683\,\textcolor{red}{$\uparrow$} \\

\bottomrule
\end{tabular}}
\end{table*}

\begin{table*}[t]
\centering
\caption{Token-level coldness taxonomy for unseen target items on each variant. Arrows indicate relative change compared to TIGER (\textcolor{red}{$\uparrow$},\textcolor{blue}{$\downarrow$}: >100\%, \textcolor{red!40}{$\uparrow$},\textcolor{blue!40}{$\downarrow$}: $\le$100\%).}
\label{tab:variant_token_support_full}
\resizebox{0.9\textwidth}{!}{
\begin{tabular}{l l cc cc cc cc}
\toprule
\multirow{2}{*}{\textbf{Dataset}}
& \multirow{2}{*}{\textbf{Coldness category}}
& \multicolumn{2}{c}{\textbf{TIGER}}
& \multicolumn{2}{c}{\textbf{TIGER-SID}}
& \multicolumn{2}{c}{\textbf{TIGER-Scorer}}
& \multicolumn{2}{c}{\textbf{TIGER-Edge}} \\
\cmidrule(lr){3-4}\cmidrule(lr){5-6}\cmidrule(lr){7-8}\cmidrule(lr){9-10}
&
& Recall@5 & Recall@20
& Recall@5 & Recall@20
& Recall@5 & Recall@20
& Recall@5 & Recall@20 \\
\midrule

\multirow{5}{*}{Beauty}
& all-token-seen                
& \z & \z 
& 0.00001\,\textcolor{red}{$\uparrow$} & 0.00001\,\textcolor{red}{$\uparrow$} 
& 0.00034\,\textcolor{red}{$\uparrow$} & 0.00253\,\textcolor{red}{$\uparrow$} 
& \z & \z \\

& any-token-unseen              
& \z & \z 
& \z & \z 
& 0.00027\,\textcolor{red}{$\uparrow$} & 0.00077\,\textcolor{red}{$\uparrow$} 
& \z & \z \\

& prefix-seen-$c_1$             
& \z & \z 
& 0.00001\,\textcolor{red}{$\uparrow$} & 0.00001\,\textcolor{red}{$\uparrow$} 
& 0.00036\,\textcolor{red}{$\uparrow$} & 0.00197\,\textcolor{red}{$\uparrow$} 
& \z & \z \\

& prefix-seen-$c_1,c_2$         
& \z & \z 
& 0.00002\,\textcolor{red}{$\uparrow$} & 0.00002\,\textcolor{red}{$\uparrow$} 
& 0.00228\,\textcolor{red}{$\uparrow$} & 0.01082\,\textcolor{red}{$\uparrow$} 
& \z & \z \\

& prefix-seen-$c_1,c_2,c_3$     
& \z & \z 
& 0.00004\,\textcolor{red}{$\uparrow$} & 0.00004\,\textcolor{red}{$\uparrow$} 
& \z & 0.18627\,\textcolor{red}{$\uparrow$} 
& \z & \z \\

\midrule

\multirow{5}{*}{Sports}
& all-token-seen                
& 0.00001 & 0.00006 
& \z\,\textcolor{blue!40}{$\downarrow$} & 0.00019\,\textcolor{red}{$\uparrow$} 
& 0.00014\,\textcolor{red}{$\uparrow$} & 0.00210\,\textcolor{red}{$\uparrow$} 
& 0.00004\,\textcolor{red!40}{$\uparrow$} & 0.00024\,\textcolor{red!40}{$\uparrow$} \\

& any-token-unseen              
& \z & \z 
& \z & \z 
& 0.00254\,\textcolor{red}{$\uparrow$} & 0.00274\,\textcolor{red}{$\uparrow$} 
& \z & \z \\

& prefix-seen-$c_1$             
& 0.00001 & 0.00004 
& \z\,\textcolor{blue!40}{$\downarrow$} & 0.00017\,\textcolor{red}{$\uparrow$} 
& 0.01343\,\textcolor{red}{$\uparrow$} & 0.01441\,\textcolor{red}{$\uparrow$} 
& 0.00003\,\textcolor{red!40}{$\uparrow$} & 0.00016\,\textcolor{red!40}{$\uparrow$} \\

& prefix-seen-$c_1,c_2$         
& 0.00002 & 0.00008 
& \z\,\textcolor{blue!40}{$\downarrow$} & 0.00035\,\textcolor{red}{$\uparrow$} 
& 0.16681\,\textcolor{red}{$\uparrow$} & 0.16971\,\textcolor{red}{$\uparrow$} 
& 0.00006\,\textcolor{red!40}{$\uparrow$} & 0.00036\,\textcolor{red!40}{$\uparrow$} \\

& prefix-seen-$c_1,c_2,c_3$     
& 0.00336 & 0.01007 
& \z\,\textcolor{blue!40}{$\downarrow$} & 0.00063\,\textcolor{blue!40}{$\downarrow$} 
& \z\,\textcolor{blue!40}{$\downarrow$} & \z\,\textcolor{blue!40}{$\downarrow$} 
& 0.00671\,\textcolor{red}{$\uparrow$} & 0.02349\,\textcolor{red}{$\uparrow$} \\

\midrule

\multirow{5}{*}{Toys}
& all-token-seen                
& \z & \z 
& \z & \z 
& 0.00132\,\textcolor{red}{$\uparrow$} & 0.00516\,\textcolor{red}{$\uparrow$} 
& \z & \z \\

& any-token-unseen              
& \z & \z 
& \z & \z 
& \z & 0.00198\,\textcolor{red}{$\uparrow$} 
& \z & \z \\

& prefix-seen-$c_1$             
& \z & \z 
& \z & \z 
& 0.00062\,\textcolor{red}{$\uparrow$} & 0.00359\,\textcolor{red}{$\uparrow$} 
& \z & \z \\

& prefix-seen-$c_1,c_2$         
& \z & \z 
& \z & \z 
& 0.00170\,\textcolor{red}{$\uparrow$} & 0.00657\,\textcolor{red}{$\uparrow$} 
& \z & \z \\

& prefix-seen-$c_1,c_2,c_3$     
& \z & \z 
& \z & \z 
& \z & \z 
& \z & \z \\

\midrule

\multirow{5}{*}{Sephora}
& all-token-seen                
& \z & \z 
& \z & \z 
& \z & 0.00003\,\textcolor{red}{$\uparrow$} 
& 0.00006\,\textcolor{red}{$\uparrow$} & 0.00242\,\textcolor{red}{$\uparrow$} \\

& any-token-unseen              
& \z & \z 
& \z & \z 
& \z & \z
& 0.00005\,\textcolor{red}{$\uparrow$} & 0.00038\,\textcolor{red}{$\uparrow$} \\

& prefix-seen-$c_1$             
& \z & \z 
& \z & \z 
& \z & 0.00002\,\textcolor{red}{$\uparrow$} 
& 0.00007\,\textcolor{red}{$\uparrow$} & 0.00105\,\textcolor{red}{$\uparrow$} \\

& prefix-seen-$c_1,c_2$         
& \z & \z 
& \z & \z 
& \z & 0.00002\,\textcolor{red}{$\uparrow$} 
& 0.00004\,\textcolor{red}{$\uparrow$} & 0.00167\,\textcolor{red}{$\uparrow$} \\

& prefix-seen-$c_1,c_2,c_3$     
& \z & \z 
& \z & \z 
& \z & \z 
& \z & \z \\

\midrule

\multirow{5}{*}{WikiLife}
& all-token-seen                
& \z & 0.00045 
& \z & 0.00008\,\textcolor{blue!40}{$\downarrow$} 
& 0.00029\,\textcolor{red}{$\uparrow$} & 0.00230\,\textcolor{red}{$\uparrow$} 
& \z & \z \\

& any-token-unseen              
& \z & \z 
& \z & \z 
& \z & \z 
& \z & \z \\

& prefix-seen-$c_1$             
& \z & 0.00042 
& \z & 0.00007\,\textcolor{blue!40}{$\downarrow$} 
& 0.00029\,\textcolor{red}{$\uparrow$} & 0.00218\,\textcolor{red}{$\uparrow$} 
& \z & \z \\

& prefix-seen-$c_1,c_2$         
& \z & 0.00175 
& \z & 0.00026\,\textcolor{blue!40}{$\downarrow$} 
& 0.00047\,\textcolor{red}{$\uparrow$} & 0.00621\,\textcolor{red}{$\uparrow$} 
& \z & \z \\

& prefix-seen-$c_1,c_2,c_3$     
& \z & \z 
& \z & 0.00086\,\textcolor{red}{$\uparrow$} 
& \z & \z 
& \z & \z \\

\midrule

\multirow{5}{*}{WeiboDaily}
& all-token-seen                
& 0.00006 & 0.00092 
& 0.00169\,\textcolor{red}{$\uparrow$} & 0.00433\,\textcolor{red}{$\uparrow$} 
& 0.00029\,\textcolor{red!40}{$\uparrow$} & 0.00069\,\textcolor{blue!40}{$\downarrow$} 
& 0.00784\,\textcolor{red}{$\uparrow$} & 0.01604\,\textcolor{red}{$\uparrow$} \\

& any-token-unseen              
& -- & -- 
& -- & -- 
& -- & -- 
& -- & -- \\

& prefix-seen-$c_1$             
& 0.00006 & 0.00092 
& 0.00169\,\textcolor{red}{$\uparrow$} & 0.00433\,\textcolor{red}{$\uparrow$} 
& 0.00029\,\textcolor{red!40}{$\uparrow$} & 0.00069\,\textcolor{blue!40}{$\downarrow$} 
& 0.00784\,\textcolor{red}{$\uparrow$} & 0.01604\,\textcolor{red}{$\uparrow$} \\

& prefix-seen-$c_1,c_2$         
& 0.00009 & 0.00138 
& 0.00210\,\textcolor{red}{$\uparrow$} & 0.00540\,\textcolor{red}{$\uparrow$} 
& 0.00017\,\textcolor{red!40}{$\uparrow$} & 0.00065\,\textcolor{blue!40}{$\downarrow$} 
& 0.01015\,\textcolor{red}{$\uparrow$} & 0.02241\,\textcolor{red}{$\uparrow$} \\

& prefix-seen-$c_1,c_2,c_3$     
& 0.00080 & 0.01386 
& 0.00295\,\textcolor{red}{$\uparrow$} & 0.00761\,\textcolor{red}{$\uparrow$} 
& \z\,\textcolor{blue!40}{$\downarrow$} & \z\,\textcolor{blue!40}{$\downarrow$} 
& 0.03332\,\textcolor{red}{$\uparrow$} & 0.09068\,\textcolor{red}{$\uparrow$} \\

\midrule

\multirow{5}{*}{WeiboTech}
& all-token-seen                
& \z & \z 
& 0.00021\,\textcolor{red}{$\uparrow$} & 0.00057\,\textcolor{red}{$\uparrow$} 
& 0.00040\,\textcolor{red}{$\uparrow$} & 0.00145\,\textcolor{red}{$\uparrow$} 
& 0.01058\,\textcolor{red}{$\uparrow$} & 0.02069\,\textcolor{red}{$\uparrow$} \\

& any-token-unseen              
& \z & \z 
& \z & \z 
& \z & \z 
& \z & \z \\

& prefix-seen-$c_1$             
& \z & \z 
& 0.00021\,\textcolor{red}{$\uparrow$} & 0.00057\,\textcolor{red}{$\uparrow$} 
& 0.00040\,\textcolor{red}{$\uparrow$} & 0.00145\,\textcolor{red}{$\uparrow$} 
& 0.01058\,\textcolor{red}{$\uparrow$} & 0.02069\,\textcolor{red}{$\uparrow$} \\

& prefix-seen-$c_1,c_2$         
& \z & \z 
& 0.00028\,\textcolor{red}{$\uparrow$} & 0.00082\,\textcolor{red}{$\uparrow$} 
& 0.00049\,\textcolor{red}{$\uparrow$} & 0.00156\,\textcolor{red}{$\uparrow$} 
& 0.01345\,\textcolor{red}{$\uparrow$} & 0.02541\,\textcolor{red}{$\uparrow$} \\

& prefix-seen-$c_1,c_2,c_3$     
& \z & \z 
& 0.00038\,\textcolor{red}{$\uparrow$} & 0.00114\,\textcolor{red}{$\uparrow$} 
& \z & \z
& 0.01910\,\textcolor{red}{$\uparrow$} & 0.02390\,\textcolor{red}{$\uparrow$} \\

\bottomrule
\end{tabular}}
\end{table*}

\end{document}